\pdfoutput=1

\documentclass[11pt]{article}

\usepackage[]{EMNLP2022}
\usepackage{times}
\usepackage{latexsym}

\usepackage[T1]{fontenc}

\usepackage[utf8]{inputenc}

\usepackage{microtype}

\usepackage{inconsolata}

\usepackage{multirow}
\usepackage{array}
\usepackage{graphicx}
\usepackage{epstopdf}
\usepackage{amsmath}
\usepackage{amssymb}
\usepackage{ulem}

\usepackage[inline]{enumitem}
\usepackage{romannum}
\AtBeginDocument{\pagenumbering{arabic}}

\usepackage{float}

\usepackage{mathtools}

\usepackage{cleveref}
\usepackage{hyperref}

\usepackage{longtable}

\usepackage{algorithm}
\usepackage{algorithmic}

\newcommand{\tabincell}[2]{
\begin{tabular}{@{}#1@{}}#2\end{tabular}
}

\usepackage{color}

\definecolor{lgreen}{RGB}{0,204,102}
\definecolor{lred}{RGB}{255,51,51}
\definecolor{dred}{RGB}{173,0,0}
\definecolor{blue}{RGB}{102,178,255}
\definecolor{purple}{RGB}{204,0,204}

\definecolor{syellow}{RGB}{204,204,0}
\definecolor{sgreen}{RGB}{0,153,77}
\definecolor{sblue}{RGB}{0,102,204}
\definecolor{sred}{RGB}{204,0,0}
\definecolor{spurple}{RGB}{204,0,204}
\definecolor{smediumvioletred}{RGB}{199,21,133}
\definecolor{Intersection}{RGB}{255,51,51}
\definecolor{Optimal}{RGB}{153,0,153}
\definecolor{HoneyOrange}{RGB}{255,179,102}
\definecolor{darkred}{RGB}{139,0,0}

\title{A Distributional Lens for Multi-Aspect Controllable Text Generation}

\author{Yuxuan Gu$^{\dag}$,Xiaocheng Feng$^{\dag \ddag}$,Sicheng Ma$^{\dag}$,Lingyuan Zhang$^{\dag}$,Heng Gong$^{\dag}$, Bing Qin$^{\dag \ddag}$\\
  $^{\dag}$Harbin Institute of Technology\quad \quad \quad $^\ddag$ Peng Cheng Laboratory\\
  \texttt{\{yxgu,xcfeng,scma,lyzhang,hgong,bqin\}@ir.hit.edu.cn} \\}

\begin{document}
\maketitle
\begin{abstract}

Multi-aspect controllable text generation is a more challenging and practical task than single-aspect control.
Existing methods achieve complex multi-aspect control by fusing multiple controllers learned from single-aspect, but suffer from attribute degeneration caused by the mutual interference of these controllers.
To address this, we provide observations on attribute fusion from a distributional perspective and propose to directly search for the intersection areas of multiple attribute distributions as their combination for generation.
Our method first estimates the attribute space with an autoencoder structure. 
Afterward, we iteratively approach the intersections by jointly minimizing distances to points representing different attributes.
Finally, we map them to attribute-relevant sentences with a prefix-tuning-based decoder. Experiments on the three-aspect control task, including sentiment, topic, and detoxification aspects, reveal that our method outperforms several strong baselines on attribute relevance and text quality and achieves the SOTA. Further analysis also supplies some explanatory support for the effectiveness of our approach\footnote{Our dataset and code are available at: \url{https://github.com/HappyGu0524/MultiControl}}.
\end{abstract}

\section{Introduction}
Controllable text generation is a challenging task in natural language generation, which aims to generate fluent text with desired attributes.
Pilot studies attempt single-aspect control by directly finetuning a conditional model \cite{ziegler2019finetuning, keskarCTRL2019}, or turn to methods with language models fixed \cite{Dathathri2020Plug} due to the high cost of large-scale pre-trained language models \cite{NEURIPS2020_1457c0d6, zhang2022opt}.

Recent works focus on a more practical setting, multi-aspect\footnote{For example, \textit{positive} is an attribute from sentiment aspect while \textit{sports} is an attribute from topic aspect.} controllable text generation, with existing approaches mainly divided into three technical routes: weighted decoding \cite{Dathathri2020Plug, krause-etal-2021-gedi-generative}, multi-objective optimization \cite{kumar2021controlled, mireshghallah-etal-2022-mix}, and prefix-tuning \cite{qian-etal-2022-controllable}, which explore ways to combine controllers learned from single-aspect and apply them to a fixed language model yet suffering from attribute degeneration caused by the mutual interference of controllers.

\begin{figure}[t]
  \centering
  \includegraphics[width=\columnwidth]{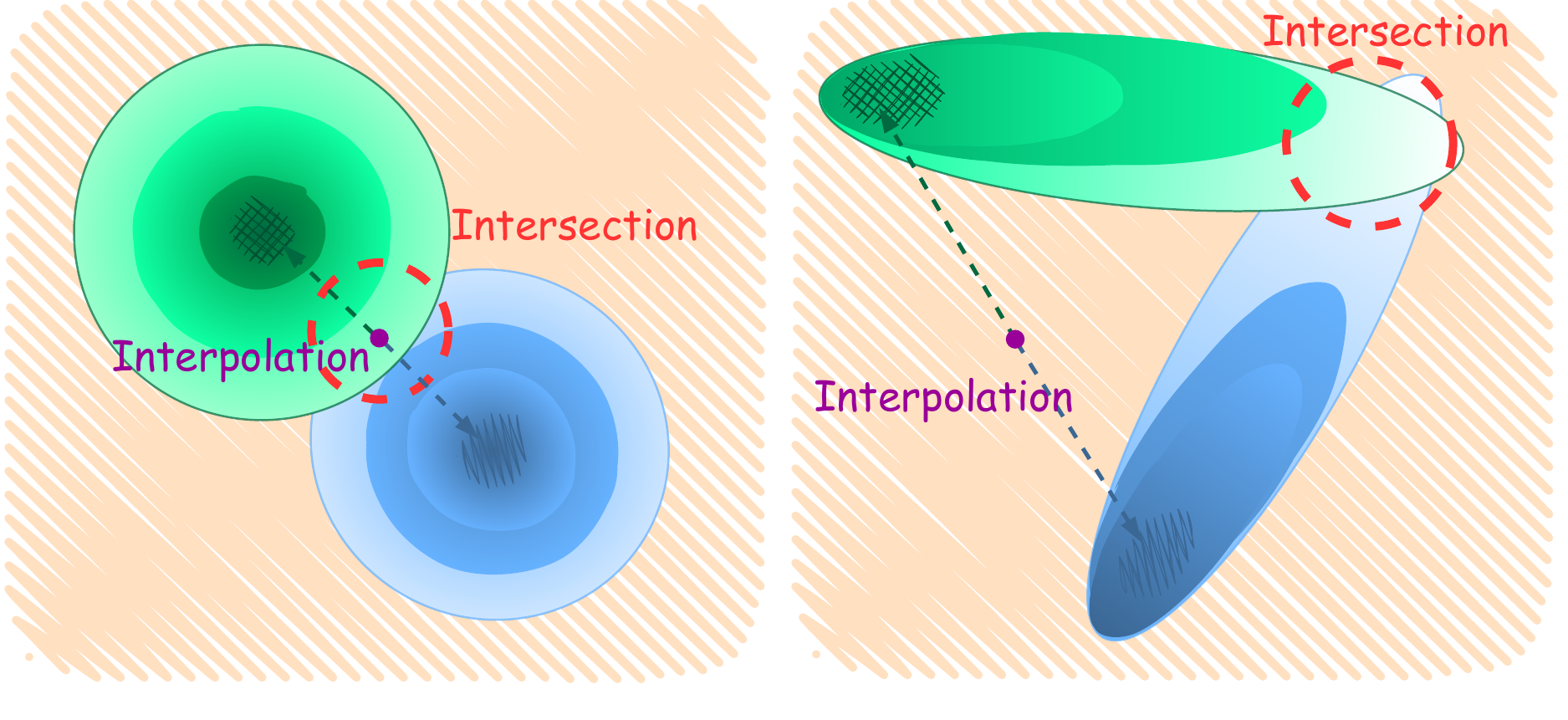}
  \vspace{-0.8cm}
  \caption{
  Probability space of attributes. \textcolor{HoneyOrange}{Orange} background denotes the estimated distribution over natural language. \textcolor{blue}{Blue} and \textcolor{lgreen}{green} areas represent distributions over sentences containing attributes from two different aspects, respectively. The darker region means a higher probability in the space. The shaded are distributional centers, the areas with the highest probability density.
  }
  \label{fig:1}
\end{figure}

We provide a distributional perspective to observe and alleviate this problem.
In the current text generation paradigm, a language model forms an estimated distribution over sentences with training data amounted to sampling from natural language distribution \cite{NEURIPS2021_260c2432}.
For single-aspect control, these methods train a classifier or a prefix for each attribute independently, which is regarded as appraising a center of distribution over attribute-relevant sentences, before biasing the language model's distribution to this center.
Correspondingly, when generalizing to multi-aspect control, their fusion strategy is directly obtaining interpolation or average of these centers, which may be too straightforward.
As shown in Figure \ref{fig:1}, the \textbf{\textcolor{Optimal}{interpolation}} point denotes the position they acquired after combining multiple centers in the probability space. And the \textbf{\textcolor{Intersection}{intersection}} represents where oracle sentences that simultaneously satisfy multiple attributes lie.
In the left part of Figure \ref{fig:1}, when distributions of attributes is symmetric\footnote{We plot distributions of attributes in \S \ref{attributes}.}, the interpolation point is indeed within the intersection area.
However, there could be a mismatch between the interpolation point and intersection. For example, as illustrated in the right part of Figure \ref{fig:1}, two skewed distributions intersect on the tails, leaving the interpolation point out of the intersection area and thus making it lack the ability to express all desired attributes together.

In this paper, different from approximating the intersection area with the interpolation point, we propose a strategy for directly acquiring the intersection.
We first deploy an autoencoder structure to map attribute-relevant sentences to latent representations constituting an estimated attribute space.
With our specially designed constraints, this space can model relationships among attributes.
Afterward, we provide an effective intersection searching algorithm that can walk around the long tail regions in distributions of all desired attributes and iteratively find where they combine more tightly.
Finally, we utilize a prefix-tuning-based decoder to construct sentences from the searched intersection. 

We experiment on three-aspect control with two attributes from the sentiment aspect, four from the topic, and one from detoxification, with datasets IMDb movie reviews \cite{maas-etal-2011-learning}, AGNews \cite{NIPS2015_250cf8b5}, and Jigsaw Toxic Comment Classification Challenge Dataset, respectively. We evaluate the relevance of each attribute independently and calculate their average as the final relevance metric. Besides, we assess the text quality with perplexity and distinctness concerning fluency and diversity.
Results show that our method can significantly outperform strong baseline models on multi-aspect control. Furthermore, we find out in our analytical experiments that our intuitive assumptions fit well with our observation. The main contributions are as follows:
\begin{itemize}[noitemsep,topsep=0pt,parsep=0pt]
    \item We propose a distributional perspective that models multi-aspect control more practically.
    \item We provide a method that directly searches for intersections in the attribute space and generates sentences with desired attributes.
    \item We experimentally reveal the effectiveness of our method on multi-aspect control compared to strong baselines and achieve the SOTA.
\end{itemize}

\section{Related Work}
Variational autoencoders are often used for controllable text generation in early work \cite{10.5555/3305381.3305545, duan-etal-2020-pre, mai-etal-2020-plug} where they spend a lot of effort into improving text fluency. 
The prosperity of large-scale pre-trained language models  \cite{radford2019language} provides more exploration directions for attribute control such as fine-tuning \cite{ficler-goldberg-2017-controlling, ziegler2019finetuning, keskarCTRL2019}. Recent work has made gratifying progress on single-aspect control \cite{krause-etal-2021-gedi-generative}, leading studies gradually turn to a more difficult task, multi-aspect control, including the following three main approaches.

\paragraph{Weighted Decoding}
As the scale of language models increases rapidly, weighted decoding \cite{Dathathri2020Plug, krause-etal-2021-gedi-generative, yang-klein-2021-fudge, liu-etal-2021-dexperts, gu-etal-2022-improving} becomes a simple and practical choice.
It is a framework that decomposes the probability of sentences conditioned on attributes into a language model and a classifier with the bayesian rule directly at decoding time. 
When handling multi-aspect control, it can be easily generalized by interpolating classifiers \cite{lin-riedl-2021-plug}.

\paragraph{Multi-Objective Optimization}
Controllable text generation task is naturally a multi-objective optimization problem when regarding its decoding process as an optimization objective.
Some approaches, such as DGC \cite{khalifa2020distributional}, Mix\&Match \cite{mireshghallah-etal-2022-mix}, and COLD Decoding \cite{qin2022cold}, adopt Energy-based Models \cite{lecun2006tutorial} to blend multiple objectives.
Others like MUCOCO \cite{kumar2021controlled} convert the optimization objectives of multi-aspect control to inequality constraints and thereby apply the lagrange multiplier method for this constrained optimization problem.

\paragraph{Prefix-Tuning}
GPT-3 \cite{brown2020language} provides a new paradigm named prompt-based learning \cite{liu2021pre}, which is able to perform few-shot learning on downstream tasks. Prefix-Tuning \cite{li-liang-2021-prefix} leverages the learned lightweight prompts to trigger the conditional generation capability of the language model.
Applying Prefix-Tuning to multi-aspect controllable text generation \cite{yu-etal-2021-attribute-alignment, qian-etal-2022-controllable, carlsson-etal-2022-fine, yang2022tailor} can be regarded as optimizing on multi-objective implicitly.

\begin{figure*}[ht]
  \centering
  \vspace{-0.5cm}
  \resizebox{2\columnwidth}{!}{
	\includegraphics[scale=0.55]{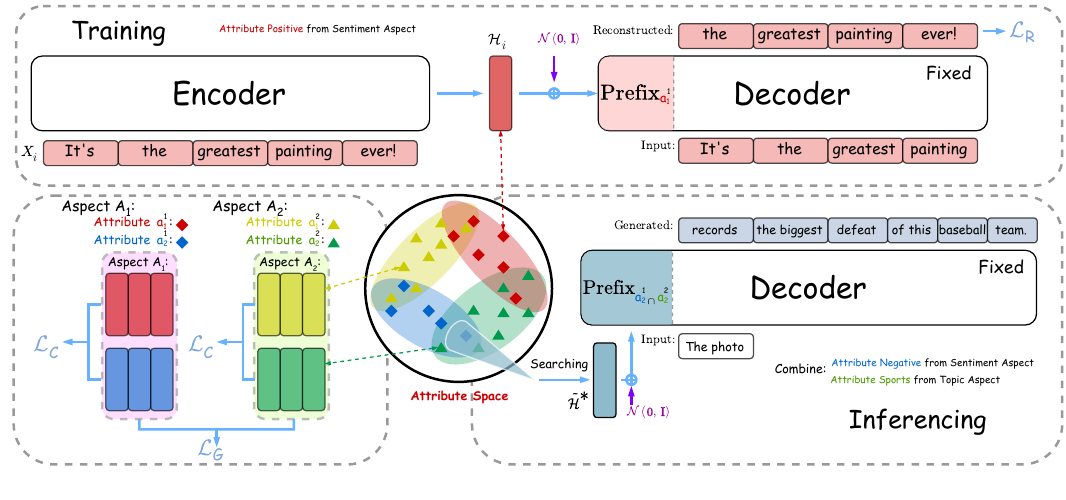}
  }
  \vspace{-0.5cm}
  \caption{An overview of our method. \textbf{Top}: Illustration of our autoencoder structure with prefix-tuning deployed on the fixed decoder, where latent representations $\mathcal{H}_i$ constitute an estimated attribute space. \textbf{Bottom Left}: Illustration of attribute classification loss $\mathcal{L}_C$ and aspect gap loss $\mathcal{L}_G$ attached to the attribute space. \textbf{Bottom Right}: Inferencing stage with prefix mapped from the intersection of attributes.}
  \label{fig:2}
\end{figure*}

\section{Methodology}

In this section, we first introduce the motivation and overall process of our method, after which we describe each module in detail.

\subsection{Overview}
As illustrated in Figure \ref{fig:2}, our method mainly revolves around the attribute space including estimating the attribute space, searching for intersections, and mapping intersections to sentences.

Firstly, we aim to construct an attribute space using sampled sentences to estimate the real space as accurately as possible. We employ an autoencoder structure with the latent representations denoting points that constitute our estimated attribute space. To ensure that our estimated space reliably models the attributes, such as their probability distributions and relationships between different attributes, we further attach three constraints to the representation.
\begin{enumerate*}[label=(\Roman*)]
    \item \textbf{Reconstruction Loss} $\mathcal{L}_R$ aims to bridge the gap between points in attribute space and natural attribute-relevant sentences, which is recovering attributes reflected by contents.
    \item \textbf{Attribute Classification Loss}  $\mathcal{L}_C$ forces the encoder to focus more on capturing attributes by distinguishing points of different attributes from the same aspect.
    \item \textbf{Aspect Gap Loss} $\mathcal{L}_G$ %
    penalizes the discrepancy of aspects, which is caused by the domain gap among different data sources for different aspects.
    Inspired by the feature alignment \cite{10.1145/1772690.1772767}, we minimize the distances between distributional centers of each two aspects.
\end{enumerate*}

The second step aims to search for an intersection area of desired attributes. If the intersection area exists, a point in the area satisfies that neighbor points appearing in a tiny surrounding region should cover all required attributes. Inspired by this neighborhood ideology, we design an algorithm that iteratively approaches an area where these attributes bind more tightly.
The third step maps our searched intersection to a Prefix that activates the language model to generate attribute-relevant sentences. To make the language model less sensitive to slight variations, we sample a perturbation vector from a multivariate gaussian distribution.

\subsection{Estimating Attribute Space}

Given $|\mathbf{A}|$ aspects $\mathbf{A} = \left\{A_1,\cdots,A_{|\mathbf{A}|}\right\}$ with each comprising $|A_t|$ attributes $\left\{a_1^t,\cdots,a_{|A_t|}^t\right\}$, $I_\tau^t$ is an index set representing the identifiers of all sentences with attribute $a^t_\tau$ in the training data. We have $I^t = \bigcup\limits_{\tau=1}^{|A_t|} I_\tau^t, I = \bigcup\limits_{t=1}^{|\mathbf{A}|} I^t$, where $I^t$ is the indices of all sentences with any attribute in aspect $A_t$ and $I$ is the indices of the entire training data.
We encode sentences $\{X_i\}$ from all aspects $\mathbf{A}$ to representations $\{\mathcal{H}_i\}$ with unified mapping parameters $\phi$: $\mathcal{H}_i = \text{Encode}_\phi(X_i)$, where $i \in I$.

\paragraph{Reconstruction Loss $\mathcal{L}_R$}
As in the top of Figure \ref{fig:2}, $\mathcal{L}_R$ is computed in the same way as the autoregressive loss of pre-trained language model $p_{\text{LM}}$:
\begin{equation}
  \setlength{\abovedisplayskip}{6pt}
  \setlength{\belowdisplayskip}{6pt}
\begin{aligned}
\label{eq:2}
\mathcal{L}_R &= -\sum\limits_{i\in I} \log p_\text{LM}(X_i|\text{Prefix}_i)\\ 
\text{Prefix}_i &= \text{MLP}_\theta(\mathcal{H}_i + \lambda \varepsilon_i),\; \varepsilon_i \sim \mathcal{N}(\mathbf{0}, \mathbf{I}),
\end{aligned}
\end{equation}
where $X_i$ here is a sample sentence from the entire training set, i.e., $i \in I$. 
Besides, $\varepsilon_i$, with a scaling factor $\lambda$, is a perturbation vector sampled from a multivariate gaussian distribution $\mathcal{N}(\mathbf{0}, \mathbf{I})$ for robustness when reconstructing. The multi-layer perceptron $\text{MLP}_{\theta}$ will map perturbed $\mathcal{H}_i$ to $\text{Prefix}_i$ that can activate the language model to generate text with desired attributes.
It's worth noting that our primary goal is to recover attributes, which means $\mathcal{L}_R$ does not need and preferably does not converge too well while maintaining text fluency.

\paragraph{Attribute Classification Loss $\mathcal{L}_C$}
We force the encoder to focus on attributes by $\mathcal{L}_C$ in the way:
\begin{equation}
  \setlength{\abovedisplayskip}{6pt}
  \setlength{\belowdisplayskip}{6pt}
  \label{eq:3}
\begin{aligned}
\mathcal{L}_C = -\sum\limits_{t=1}^{|\mathbf{A}|}\sum\limits_{\tau=1}^{|A_t|} \sum\limits_{i \in I_\tau^t} \log p_{\pi_{t}}(a_\tau^t|\mathcal{H}_i).
\end{aligned}
\end{equation}
Given sentence representation, $p_{\pi_t}$ is a classifier that distinguish attributes $\left\{a_\tau^t\right\}$ from aspect $A_t$ with parameter $\pi_{t}$.

\paragraph{Aspect Gap Loss $\mathcal{L}_G$}
We penalize the discrepancy between distributional centers by:
\begin{equation}
  \setlength{\abovedisplayskip}{6pt}
  \setlength{\belowdisplayskip}{6pt}
  \label{eq:4}
\mathcal{L}_G = \sum\limits_{\mathclap{1\leq t_1<t_2\leq |\mathbf{A}|}}\quad\;\,\,\left\|\sum\limits_{i \in I^{t_1}} \frac{\mathcal{H}_i}{|I^{t_1}|} - \sum\limits_{j \in I^{t_2}} \frac{\mathcal{H}_j}{|I^{t_2}|}\right\|_2,
\end{equation}
which are Euler distances between every two distinct distributional centers.
When generalizing to a larger scale of aspects, it is relatively expensive to calculate averages over the entire dataset each time the model is updated.
We calculate this loss in practice using a batch-level approximation. We assign each aspect a memory unit to store the latest representation of the aspect's estimated center.
Each time processing a batch of sentences from one aspect, we take the average of their representations as the center and sum up the Euler distances to centers of other aspects in the memory, which is the estimated $\mathcal{L}_G$. Then, we update the memory unit of this aspect to the latest.

\begin{algorithm}[t]
	\renewcommand{\algorithmicrequire}{\textbf{Input:}}
	\renewcommand{\algorithmicensure}{\textbf{Output:}}
	\caption{Intersection Searching}
	\label{alg1}
	\begin{algorithmic}[1]
	    \REQUIRE $\mathcal{H}_i, i \in \bigcup\limits_{t=1}^N I^t_{\alpha_t}$ from $N$ attributes\\
	    \quad \; $\omega_{\alpha_t}$ weight of each attribute
	    \ENSURE Intersection of $N$ attributes: $\mathcal{\tilde{H}}^*$
		\STATE Initialize $M$ candidates:$\{\mathcal{\tilde{H}}^0_m\}$ 
		\STATE Iterate $S$ times
		\FOR{$s$ in $[0, S-1]$}
		    \FOR{$m$ in $[1, M]$}
		        \STATE $\mathcal{\tilde{H}}_m^{s+1}\leftarrow \mathbf{0}$
		        \FOR{$t$ in $[1, N]$}
    		        \STATE $\mathbf{H}\leftarrow\mathop{\text{Nearest}}\limits_{top K}(\mathcal{\tilde{H}}_m^s,\left\{\mathcal{H}_i, i \in I^t_{\alpha_t}\right\})$
    		        \STATE $\mathcal{\tilde{H}}_m^{s+1}\leftarrow \mathcal{\tilde{H}}_m^{s+1} +\omega_{\alpha_t} \mathop{\text{mean}}(\mathbf{H})$
		        \ENDFOR
		        \STATE $\mathcal{\tilde{H}}_m^{s+1}\leftarrow \mathcal{\tilde{H}}_m^{s+1} / \sum\limits_{t=1}^N \omega_{\alpha_t}$
		    \ENDFOR
		\ENDFOR
		\STATE $\mathcal{\tilde{H}}^*\leftarrow \text{Select}(\{\mathcal{\tilde{H}}^{S}_m\})$

	\end{algorithmic} 
\end{algorithm}

During the training stage, our loss function is:
\begin{equation}
  \setlength{\abovedisplayskip}{6pt}
  \setlength{\belowdisplayskip}{6pt}
  \label{eq:5}
  \mathcal{L} = w_1\mathcal{L}_R + w_2\mathcal{L}_C + w_3\mathcal{L}_G.
\end{equation}
It's worth noting that we only update parameters $\phi$, $\theta$, and $\left\{\pi_t\right\}$ for the encoder, the MLP layer, and the classifier heads, respectively.

\subsection{Intersection of Attributes}
Suppose there is an intersection point, denoted as $\mathcal{\tilde{H}}^*$, located within the intersection region of attributes $\left\{a^1_{\alpha_1},a^2_{\alpha_2},\cdots,a^N_{\alpha_N}\right\}$ from $N$ different aspects, where $a^t_{\alpha_t}$ is the $\alpha_t$th attribute in aspect $A_t$. 
Our algorithm \ref{alg1} approximates the $\mathcal{\tilde{H}}^*$ by iteratively approaching a most balanced point with nearest neighbors from different attributes.
First, we initialize the candidates $\{\mathcal{\tilde{H}}^0_m\}$ by randomly sampling points in the attribute space, calculating their distance to the closest point of each attribute $a^t_{\alpha_t}$, and selecting the top $M$ samples with the smallest average distance to all attributes.
At each iteration $s$, we choose the top-K\footnote{We study the practical meaning and impact of $K$ in \S \ref{sec:effectofk}.} nearest points to $\mathcal{\tilde{H}}^s_m$ for each attribute and update $\mathcal{\tilde{H}}^{s+1}_m$ using the weighted average of these points.
It is worth mentioning that $\omega_{\alpha_t}$ is the weight used to balance attributes or favor some specifically, and a negative value of $\omega_{\alpha_t}$ can even move away from a particular one.
Finally, we select the best candidate from the last iteration $S$, which is expected to be in the intersection region, i.e., a representation related to multiple attributes.

\subsection{Generation with Intersections}
As illustrated in the right bottom of Figure \ref{fig:2}, we convert the representation $\mathcal{\tilde{H}}^*$ obtained from the intersection area directly to the $\text{Prefix}$ with $\text{MLP}_\theta$ and let the language model generate multi-attributed sentence $Y$ from input $\mathcal{X}$ as:
\begin{equation}
  \setlength{\abovedisplayskip}{6pt}
  \setlength{\belowdisplayskip}{6pt}
  \label{eq:1}
\begin{aligned}
Y &= \mathrm{arg}\max\limits_y \; p_\text{LM}(y|\text{Prefix}^*;\mathcal{X})\\
\text{Prefix}^* &=\text{MLP}_{\theta}(\mathcal{\tilde{H}}^* + \lambda \varepsilon_i),\; \varepsilon_i \sim \mathcal{N}(\mathbf{0}, \mathbf{I}).
\end{aligned}
\end{equation}
When generating several attribute-relevant sentences for one attribute combination, we only need to calculate the intersection for it once.

\section{Experiment}
In this section, we demonstrate the effectiveness of our method on three-aspect control, including sentiment, topic, and detoxification.
\subsection{Multi-Aspect Control Task}
The datasets we use are the same as GeDi \cite{krause-etal-2021-gedi-generative} and Contrastive Prefix \cite{qian-etal-2022-controllable}. 
To balance the data scale across all aspects, we randomly sample 10k sentences from each dataset that is less than the number of samples GeDi uses, with each attribute equally dividing this amount.
We use the IMDb movie reviews \cite{maas-etal-2011-learning}, the AGNews dataset \cite{NIPS2015_250cf8b5}, and the Jigsaw Toxic Comment Classification Challenge Dataset\footnote{\url{https://www.kaggle.com/c/jigsaw-toxic-comment-classification-challenge/}} for sentiment, topic and detoxification aspects, respectively.

The prompts used for text generation are the same as those used in the PPLM \cite{Dathathri2020Plug}, with 20 from its bag-of-words experiment and 15 from its discriminator experiment. We experiment with $8$ combinations of the $3$ aspects with $2$ sentiments $\times$ $4$ topics $\times$ $1$ detoxification and generate $5$ completions for each combination and each prompt. Totally, each model will generate $35 \times 2 \times 4 \times 1 \times 5 = 1400$ sentences. 
It is worth noting that we do not specifically use prompts that induce the language model to generate toxic text, making detoxification easier to improve.

To measure the performance on different aspects, we compute the attribute relevance. We finetune a DeBERTa \cite{he2021deberta, he2021debertav3} classifier on the Yelp dataset \cite{NIPS2015_250cf8b5} for sentiment aspect and a classifier for topic utilizing all its remaining data not used during training. We evaluate the non-toxicity with the Google Perspective API\footnote{\url{https://www.perspectiveapi.com}}.
The final performance of a model is determined by the average of these three attribute relevance scores introduced above. 
We also use two auxiliary metrics to measure text quality.
One is perplexity calculated by GPT2-large following Contrastive Prefix \cite{qian-etal-2022-controllable}. To ensure that models are not insensitive to changes in different prefixes, we calculate the Distinctness \cite{li-etal-2016-diversity} of sentences generated from different prefixes and average the 1-gram, 2-grams, and 3-grams distinct scores for simplicity. 
Moreover, we conduct human evaluation with sentences generated by different models shuffled. Each sentence is rated by three professional evaluators for 3 attribute relevance and text fluency. Evaluators rate each item on a scale of 1 to 5, with 5 representing text highly related to the desired attribute or very fluent.

\subsection{Baselines}
\begin{table*}[ht]
  \small
  \vspace{-0.3cm}
  \centering
      \begin{tabular}{l|c|ccc|c|c}
          \hline 
          
          \hline
          \textbf{Methods} & \textbf{Average}↑ (\%) & \textbf{Sentiment}↑ (\%) & \textbf{Topic}↑ (\%) & \textbf{Detoxification}↑ (\%) & \textbf{PPL.}↓ &\textbf{Dist.}↑\\
          \hline
          \hline
          \multicolumn{6}{l}{\quad\textit{Weighted Decoding Based Methods}}\\
          \hline
          \textbf{PPLM} & 71.0 $\pm$ 21.4 & 64.7 $\pm$ 24.8 & 63.5 $\pm$ 22.7 & 84.9 $\pm$\; 6.5 & 62.6 & 62.0\\
          \hline
          \textbf{GeDi} & 81.4 $\pm$ 14.7 & 76.1 $\pm$ 17.2 & 73.8 $\pm$ 11.3 & 94.2 $\pm$\; 1.9 & 116.6 & 75.1\\
          \hline
          \hline
          \multicolumn{6}{l}{\quad\textit{Multi-Objective Optimization Based Methods}}\\
          \hline
          \textbf{MUCOCO} & 73.9 $\pm$ 24.1 & 65.0 $\pm$ 33.7 & 67.2 $\pm$ 18.3 & 89.5 $\pm$\; 3.5 & 405.6 & 49.7\\
          \hline
          \textbf{Mix\&Match} & 79.7 $\pm$ 21.8 & 73.5 $\pm$ 25.9 & 69.9 $\pm$ 21.1 & 95.8 $\pm$\; 1.9 & 63.0 & 61.8\\
          \hline
          \hline
          \multicolumn{6}{l}{\quad\textit{Prefix-Tuning Based Methods}}\\
          \hline
          \textbf{Contrastive Prefix} &&&&&\\
          \quad concatenation & 77.2 $\pm$ 18.5 & 67.3 $\pm$ 20.7 & 71.8 $\pm$ 16.5 & 92.6 $\pm$\; 2.9 & 54.6 & 39.9\\
          \quad semi-supervised & 81.3 $\pm$ 16.5 & 74.4 $\pm$ 19.6 & 76.9 $\pm$ 16.7 & 92.7 $\pm$\; 3.5 & 31.9 & 43.3\\
          
          \hline
          \textbf{Ours} & \textbf{87.4} $\pm$ 10.9 & \textbf{86.7} $\pm$ 10.5 & \textbf{84.8} $\pm$ 14.2
          & 90.7 $\pm$\; 7.4 & 28.4 & 49.5\\
          \cline{2-7}
          \quad w/o $\mathcal{L}_G$ & 80.9 $\pm$ 16.2 & 71.6 $\pm$ 11.7 & 75.9 $\pm$ 18.9 & 95.3 $\pm$\; 2.6 & 71.5 & 58.9\\
          \quad w/o $\mathcal{L}_C$ & 62.3 $\pm$ 41.8 & 49.1 $\pm$ 49.8 & 41.7 $\pm$ 36.0 & \textbf{96.0} $\pm$\; 0.1 & 473.0 & 37.0\\
          \hline

          \hline
      \end{tabular}
\caption{Automatic Results on Multi-Aspect Control. Hyperparameters and details are in \S \ref{sec:appendix3}.}
  \label{tab:1}
  \vspace{-0.2cm}
\end{table*}

\begin{enumerate*}[label=(\Roman*)]
\item \textbf{Weighted Decoding}:
\textbf{PPLM} \cite{Dathathri2020Plug} biases the language model with gradients back-propagated from trained classifiers. \textbf{GeDi} \cite{krause-etal-2021-gedi-generative} influences the decoding process with token probabilities conditioned on attributes.
\item \textbf{Multi-objective Optimization}:
\textbf{MUCOCO} \cite{kumar2021controlled} regards the decoding process as a constrained optimization problem, where the language model is the objective function and attributes are constraints.
\textbf{Mix\&Match} \cite{mireshghallah-etal-2022-mix} controls attributes with energy-based models and generates sentences by masking, sampling, and correcting.
\item \textbf{Prefix-Tuning}:
\textbf{Contrastive Prefix} \cite{qian-etal-2022-controllable} utilizes prefixes to activate the language model to generate attribute-relevant sentences by concatenation or semi-supervision.
\end{enumerate*}

\subsection{Results}
According to the automatic evaluation results in Table \ref{tab:1}, under the multi-aspect setting, we group models based on their type of methods in chronological order.
In addition, we demonstrate their standard deviations, which reflect the stability of models among different attribute combinations. 

For weighted decoding, GeDi uses more powerful classifiers than PPLM and performs better on attribute relevance, stability to different combinations, and distinctness while correspondingly worse on perplexity. 
Multi-objective optimization methods achieve a favorable performance on attribute relevance while MUCOCO explodes on perplexity due to its non-autoregressive paradigm not being suitable for generating from scratch.
Performance of semi-supervised Contrastive Prefix is similar to GeDi, except for lack of diversity.

Our method performs best on average attribute-related metrics, with at least a $7.3\%$ significant improvement over existing baselines. Our advances mainly come from sentiment and topic aspects, with no less than $13.9\%$ and $10.3\%$ each.
Although our model is not the best on detoxification, it is the most balanced and stable according to the lowest standard deviation on average, $10.9$.
As a prefix-tuning-based method inducing the language model without direct modification, which is naturally good at text fluency, we perform well on perplexity and inherit the performance on diversity.

Furthermore, we conduct ablation on aspect gap loss $\mathcal{L}_G$ and attribute classification loss $\mathcal{L}_C$ separately.
On the one hand, without $\mathcal{L}_G$, we can not alleviate the bias in different training datasets, making it hard to search for the intersection areas. Since training sentences of sentiment and topic aspects are mainly non-toxic, our model focuses more on detoxification rather than struggling for the other two, leading to considerable declines on their relevance while slight improvements on detoxification. Besides, as the distance among sample points from different aspects in the attribute space increases, our model will generate sentences mapped from far more sparse areas, leading to a small decrease on fluency and a subtle increase on diversity.
On the other hand, without $\mathcal{L}_C$, our attribute space will totally collapse. The relevance of sentiment and topic drops drastically while the non-toxicity boosts because model can hardly distinguish representations of different attributes in the same aspect and focus on relatively more effortless detoxification.
Worse still, without distinct representations, our model is required to recover different sentences from similar ones, leading to oscillation in training and hardly generating complete text when inferencing.

Results of human evaluation are in Table \ref{tab:2}, with inter-annotator agreement being $0.36$ in Fleiss’ $\kappa$.
We evaluate GeDi, Contrastive Prefix, and our method and observe that the results are consistent with the automatic ones on sentiment and topic relevance. 
The performance of models on detoxification is high and relatively similar, making the automatic results different from the manual ones where the annotators believe that our model does a better job than baselines. 
Since perplexity is relatively unreliable, the manually measured fluency of GeDi is much better than that of the Contrastive Prefix. And our method achieves the best fluency.

\begin{table}[t]
  \small
  \centering
      \begin{tabular}{l|cccc}
          \hline 
          
          \hline
          \textbf{Methods} & \textbf{Sent.}↑ & \textbf{Topic}↑ & \textbf{Detox.}↑ &\textbf{Fluency}↑\\
          \hline
          \hline
          \textbf{GeDi} & 2.96 & 2.72 & 4.59 & 3.08\\
          \hline
          \textbf{Con. Prefix} & 2.84 & 2.90 & 4.40 & 2.26\\
          \hline
          \textbf{Ours} & \textbf{3.47} & \textbf{3.39} & \textbf{4.71} & \textbf{3.69}\\
          \hline

          \hline
      \end{tabular}
\caption{Human Evaluation on Multi-Aspect Control. }
  \label{tab:2}

\end{table}

\begin{table*}[ht]
  \small
  \centering
  \vspace{-0.3cm}
    \resizebox{2\columnwidth}{!}{
      \begin{tabular}{l|cc|cccc|c}
          \hline 
          
          \hline
          \multirow{2}{*}{\textbf{Methods}} & \multicolumn{2}{c|}{\textbf{Sentiment} (\%)} &\multicolumn{4}{c|}{\textbf{Topic} (\%)} & \multirow{2}{*}{\textbf{Detox.} (\%)}\\
           & \textbf{Neg.}& \textbf{Pos.}& \textbf{World}&\textbf{Sports}& \textbf{Business}&\textbf{Sci./Tech.}&\\
          \hline
          \hline
          \multicolumn{6}{l}{\quad \textit{Weighted Decoding Based Methods}}\\
          \hline
          \textbf{GeDi} \textit{single-aspect} & 93.9& 70.7& 73.4  & 85.7 & 75.7 & 98.0 & 94.9 \\

          \hline
          \multirow{8}{*}{\textbf{GeDi}}
          & 94.7 & - & 80.0 & - & - & - & 90.6\\
          & 84.2 & - & - & 74.8 & - & - & 93.9\\
          & 94.9 & - & - & - & 75.7 & - & 96.6\\
          & 90.6 & - & - & - & - & 80.1 & 92.8\\
          & - & 53.7 & 61.4 & - & - & - & 94.4\\
          & - & 60.5 & - & 74.3 & - & - & 95.2\\
          & - & 57.6 & - & - & 54.3 & - & 95.7\\
          & - & 72.3 & - & - & - & 90.2 & 94.2\\
          \cline{2-8}
          \qquad \textit{average} & \textbf{91.1} $(- \textbf{2.8})$ & 61.0 $(-9.7)$ & 70.7 $(-2.7)$ & 74.6 $(-11.1)$ & 65.0 $(-10.7)$ & 85.2 $(-12.8)$ & \textbf{94.2} $(-\textbf{0.7})$\\

          \hline
          \hline

          \multicolumn{6}{l}{\quad \textit{Prefix-Tuning Based Methods}}\\
          \hline
          \textbf{Prefix} \textit{single-aspect} & 88.4 & 90.6 & 74.5 & 85.3 & 93.5 & 93.6 & 93.8\\
          
          \hline

          \multirow{8}{*}{\tabincell{l}{\textbf{Contrastive Prefix}\\ \quad semi-supervised}}
          & 65.5 & - & \textbf{80.6} & - & - & - & 91.8\\
          & 67.2 & - & - & \textbf{90.3} & - & - & 92.5\\
          & 56.0 & - & - & - & 79.2 & - & 92.2\\
          & 90.0 & - & - & - & - & 93.3 & 84.8\\
          & - & 93.5 & 64.8 & - & - & - & 95.1\\
          & - & 41.8 & - & 78.5 & - & - & 94.8\\
          & - & 87.4 & - & - & 41.7 & - & 95.2\\
          & - & 93.6 & - & - & - & 86.7 & 95.3\\
          \cline{2-8}
          \qquad \textit{average} & 69.7 $(-18.7)$ & 79.1 $(-11.5)$ & \textbf{72.7} $(-\textbf{1.8})$ & \textbf{84.4} $(-\textbf{0.9})$ & 60.5 $(-33.0)$ & 90.0 $(-3.6)$ & 92.7 $(-1.1)$\\
          
          \hline
          \multirow{8}{*}{\textbf{Ours}}
          & 69.7 & - & 71.7 & - & - & - & 84.1\\
          & 78.6 & - & - & 80.0 & - & - & 80.2\\
          & \textbf{99.9} & - & - & - & \textbf{96.7} & - & 96.8\\
          & 92.8 & - & - & - & - & \textbf{98.0} & 81.7\\
          & - & 80.5 & 58.0 & - & - & - & 95.1\\
          & - & 84.7 & - & 86.6 & - & - & 94.5\\
          & - & 87.6 & - & - & 91.7 & - & \textbf{98.1}\\
          & - & \textbf{99.7} & - & - & - & 96.1 & 95.4\\
          \cline{2-8}
          \qquad \textit{average} & 85.3 $(-3.1)$ & \textbf{88.1} $(-\textbf{2.5})$ & 64.9 $(-9.6)$ & 83.3 $(-2.0)$ & \textbf{94.2} $(+\textbf{0.7})$ & \textbf{96.8} $(+\textbf{3.2})$ & 90.7 $(-3.1)$\\
          
          \hline

          \hline
      \end{tabular}
    }
\caption{Detailed Results on Single-Aspect and Multi-Aspect Control. We demonstrate results on \textit{single-aspect} and \textit{average} results on multi-aspect control with their difference to \textit{single-aspect}, where other rows each represent an attribute combination. Cases are in \S \ref{sec:appendix4}. Detailed results for other baseline models and our ablations are in \S \ref{sec:appendix5}.}
  \label{tab:3}
   \vspace{-0.2cm}
\end{table*}

\section{Analysis}

\subsection{Effect of Different Attributes and their Combinations}
We illustrate the detailed results of each attribute and their combinations in Table \ref{tab:3}.
GeDi and Prefix-tuning perform differently in \textit{single-aspect} control, each with its advantages. For example, GeDi is dedicated to \textit{negative} with $93.9\%$ relevance, while Prefix-tuning is good at \textit{positive} with $90.6\%$ relevance. 
When dealing with multi-aspect control, they inherit such imbalanced characteristics, with \textit{average} relevance of $91.1\%$ and $79.1\%$, respectively. In addition, the baselines decrease correspondingly in the \textit{average} relevance of each attribute compared to \textit{single-aspect}, ranging from $0.7$ to $33.0$. 
On average,  our model outperforms other baselines on attribute metrics (Table \ref{tab:1}). In detail, our model performs competitively for most attributes compared to another prefix-tuning-based model, Contrastive Prefix. Especially, on attributes like \textit{business} and \textit{sci/tech}, our model significantly improves over another prefix-tuning-based method on multi-aspect control and can even surpass it under \textit{single-aspect} control.

In addition, correlations between attributes vary widely, as in Table \ref{tab:3}.
For example, generally, \textit{positive} fits well with \textit{non-toxic} while \textit{negative} leads to a massive drop in non-toxicity, which is consistent with the intuition that one can hardly praise people and offend them simultaneously. 
Besides, \textit{world} and \textit{business} news are often reported negatively, such as war, famine, inflation, etc., making it challenging to combine them with \textit{positive}.
When attributes are not closely correlated, which means that few natural sentences possess these attributes together, our method is more likely to capture such a rarely occurred incident and magnify their frequency.
Take \textit{business} as an example. It is effortless to achieve a fine attribute relevance when performing single-aspect control on \textit{business}, with GeDi achieving $75.7$ and Prefix obtaining $93.5$. After attaching \textit{positive} to \textit{business}, baseline models will suffer from a decline due to their weak correlation, where GeDi and Contrastive Prefix drop to $54.3$ and $41.7$, respectively. In contrast, our method can alleviate this problem by retrieving this unusual co-occurrence in the training sentences and recovering it from the attribute space, achieving a performance of $91.7$, which is close to single-aspect control.
\begin{figure}[t]
  \centering
  \includegraphics[width=\columnwidth]{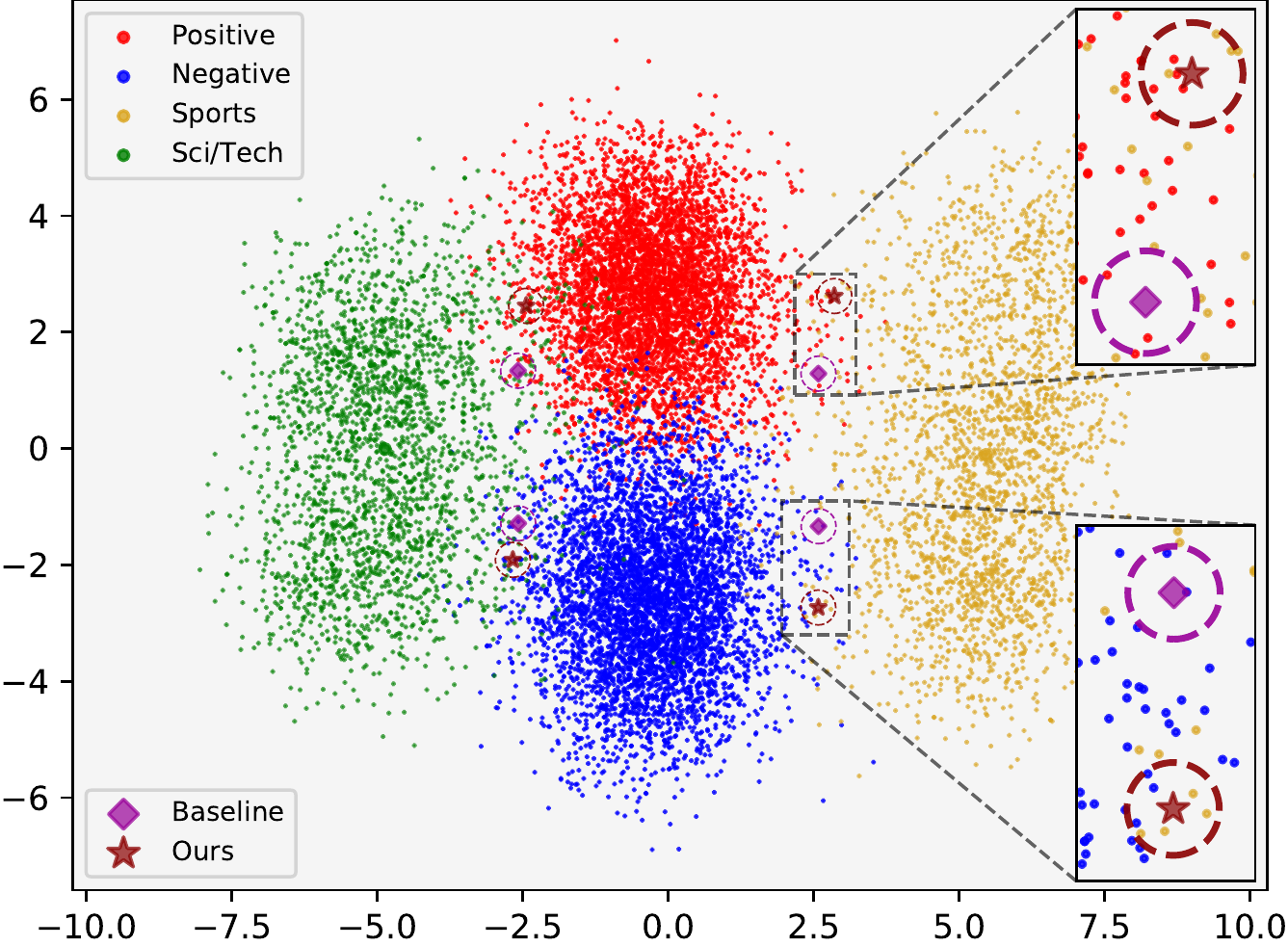}
  \caption{Projection of 4 attributes from attribute space.}
  \label{fig:3}
\end{figure}
When combining business with negative, which is a relatively common combination, there is still some decrease for baseline models. On the contrary, our method can even obtain the performance of $96.7$ that surpasses single-aspect control.

\subsection{Estimated Attribute Space}

We demonstrate part of our estimated attribute space in Figure \ref{fig:3} with four attributes: \textit{\textcolor{sred}{positive}}, \textit{\textcolor{sblue}{negative}}, \textit{\textcolor{syellow}{sports}}, and \textit{\textcolor{sgreen}{sci/tech}} from sentiment and topic aspects. 
We project the high-dimensional space to 2D with Principal Component Analysis (PCA). 
Consistent with our hypothesis, distributions of \textit{\textcolor{syellow}{sports}} and \textit{\textcolor{sgreen}{sci/tech}} are asymmetric and the intersections lie in the sparse edges of attributes' distribution.
In addition, we project the intersections searched by the \textcolor{smediumvioletred}{baseline}'s strategy and \textcolor{darkred}{ours}, respectively. For \textit{\textcolor{sred}{positive}}-\textit{\textcolor{sgreen}{sci/tech}} and \textit{\textcolor{sblue}{negative}}-\textit{\textcolor{sgreen}{sci/tech}} pairs, the combinations are relatively tight, making it easy to find intersections. However, intersection areas for \textit{\textcolor{sred}{positive}}-\textit{\textcolor{syellow}{sports}} and \textit{\textcolor{sblue}{negative}}-\textit{\textcolor{syellow}{sports}} pairs are considerably sparse. 
As shown in enlarged area, the \textcolor{smediumvioletred}{baseline} searched intersection is at the midpoint of the two distributional centers, but this location is not where the attributes intersect. On the contrary, \textcolor{darkred}{our} method can find an intersection in such a sparse region, making various points from the two different attributes appear simultaneously in its tiny surrounding area. 
It worth noting that \textit{positive} and \textit{negative} appear to intersect in this projection because they are close in the high-dimensional space. But there is actually no intersection if only projecting these two attributes in \S \ref{sec:pos_neg}. 

\subsection{Effect of $K$}
\label{sec:effectofk}
\begin{table}[t]
  \small
  \setlength{\abovecaptionskip}{0.2cm}
  \vspace{-0.3cm}
  \centering
      \begin{tabular}{r|c|ccc}
          \hline 
          
          \hline
          $\textbf{K}$ &\textbf{Avg.}↑ & \textbf{Sent.}↑ & \textbf{Topic}↑ & \textbf{DeTox.}↑\\
          \hline
          5000 & \textcolor{sblue}{75.5} & 70.5 & 67.9 & 88.2\\
          4000 & \textcolor{sblue}{77.6} & 72.9 & 71.4 & 88.4\\
          3000 & \textcolor{sblue}{78.7} & 72.4 & 74.7 & 88.9\\
          2000 & \textcolor{sblue}{79.1} & 72.6 & 75.9 & 88.7\\
          1500 & \textcolor{sblue}{79.9} & 73.6 & 77.1 & 89.0\\
          1000 & \textcolor{sblue}{80.7} & 75.7 & 77.2 & 89.1\\
          800  & \textcolor{sblue}{82.9} & 79.3 & 79.2 & 90.3\\
          500  & \textcolor{sblue}{85.2} & 83.5 & 81.5 & 90.5\\
          300  & \textcolor{sblue}{85.7} & 84.1 & 83.2 & 89.7\\
          200  & \textcolor{sred}{\textbf{87.4}} & \textbf{86.7} & \textbf{84.8} & 90.7\\
          150  & \textcolor{sgreen}{84.0} & 79.2 & 84.3 & 88.4\\
          100  & \textcolor{sgreen}{83.9} & 78.7 & 83.6 & 89.5\\
          50   & \textcolor{sgreen}{82.2} & 78.4 & 78.5 & 89.6\\
          20   & \textcolor{sgreen}{80.9} & 77.8 & 73.1 & 91.7\\
          10   & \textcolor{sgreen}{80.8} & 79.6 & 71.5 & 91.2\\
          5    & \textcolor{sblue}{81.4} & 82.9 & 69.3 & 92.1\\
          3    & \textcolor{lred}{85.0} & 86.1 & 77.7 & 91.1\\
          1    & \textcolor{sgreen}{78.8} & 63.1 & 80.9 & \textbf{92.4}\\
          \hline

          \hline
      \end{tabular}
\caption{Results that vary with $K$.}
\vspace{-0.2cm}
  \label{tab:k_analysis}
\end{table}
We analyze the variation of $K$ in the intersection searching algorithm and demonstrate the results in Table \ref{tab:k_analysis}. 
Our model reaches a critical point when $K$ is 200, and the performance is optimal this time. 
On the one hand, as the value of $K$ gradually increases, our method pays less attention to regions where samples are fewer while attributes combine more tightly, and the performance decreases accordingly.
When $K$ reaches 5k, our method degenerates into a plain prefix-tuning model, which treats intersection as the midpoint of distributional centers. Its performance is similar and slightly inferior to the concatenation version of Contrastive Prefix in Table \ref{tab:1}. 
On the other hand, smaller $K$ leads to suboptimal performance since the effect of noise becomes non-negligible in training data.
When $K$ is less than $10$, our model will be very unstable.

\subsection{Distribution of Attributes}
\label{attributes}
\begin{figure}[t]
  \centering
  \vspace{-0.3cm}
  \includegraphics[width=\columnwidth]{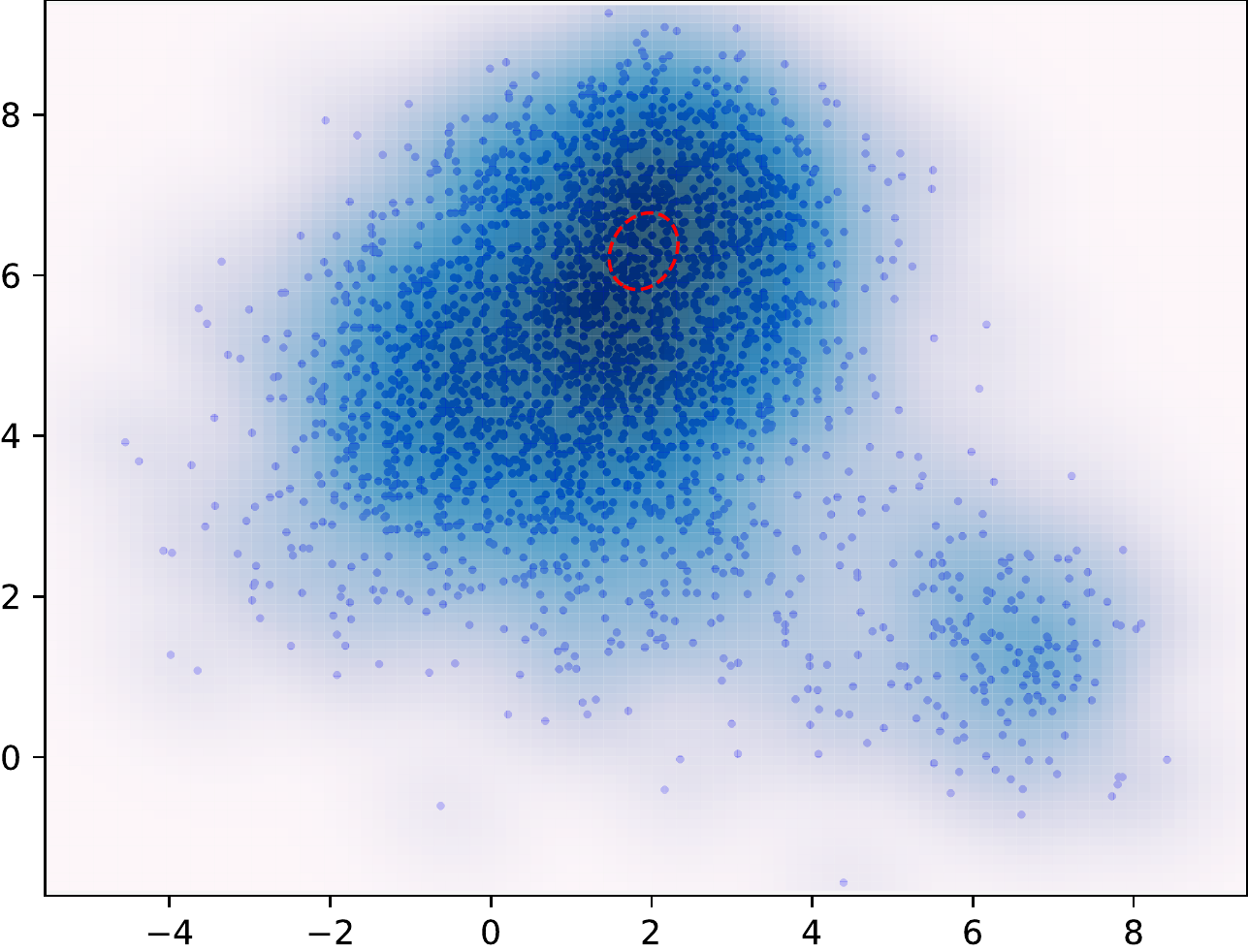}
  \caption{Distribution of attribute World from Topic.}
  \label{fig:4}
  \vspace{-0.2cm}
\end{figure}
We project sample points to 2D by PCA, with each attribute projected independently. As in Figure \ref{fig:4}, we display a scatterplot of World and conduct a Gaussian kernel density estimation to visualize its probability distribution. The darker area denotes a higher probability, where more representation points of oracle sentences gather. And the region annotated by a red ellipse is the estimated distributional center. As in the plot, the distribution of World is significantly asymmetric as the center lies in the top part, with the bottom being a sparse long tail. In addition, the distribution is even non-convex with an isolated cluster in the lower right corner. This observation supports our hypothesis that the practical distributions of attributes are far more complex than symmetric distributions such as Gaussian distribution. Besides, we plot the distribution of other attributes in the \S \ref{sec:appendix1}.

\section{Discussion on Distributional Lens}
Pilot work such as DGC \cite{khalifa2020distributional} estimates the language distribution with an energy-based model and optimizes this distribution to satisfy constraints by approaching the constraints manifold. Recent distributional approaches like COLD Decoding \cite{qin2022cold} and MuCoLa \cite{kumar2022constrained} take the language and attribute distribution in the same space so as to sample attribute-related sentences with Langevin Dynamics. Concurrent work on the image side, PromptGen \cite{wu2022generative}, simulates the complex distribution of images relevant to target attributes using a deep generative model. However, as a consensual hypothesis in manifold learning, the pre-trained language model estimates a low-dimensional manifold of language in a high-dimensional embedding space, which means most points in the embedding space are not probabilistically modeled by the language model. We believe that placing too much trust in the distributional modeling ability of language models is not a good choice. Our method attempts to depict the attribute space with discrete sample points of attributed sentences and make these discrete points, along with their coverage areas, compose the support set of our estimated distribution.

\section{Conclusion}
In this work, we present a distributional perspective for the multi-aspect controllable text generation with experimental results confirming the superiority of our model. Further observations on the 2D projection of the estimated attribute space show that our hypothesis about the attribute space is more feasible. In the future, we can explore the correlation between different attribute combinations for more fine-grained control and capture the bias in datasets to eliminate or utilize it.

\section*{Limitations}
Our method has a certain dependence on the data since we need to estimate an attribute space. Therefore, it is difficult for our method to perform well in the setting of few-shot learning. However, this disadvantage is not that severe, because we only need single-aspect data, which is relatively sufficient in style transfer tasks. Another dependence of our method on data is that it is somewhat sensitive to biases in the data. When the semantic divergence of different aspects in training data is too large, our aspect gap loss, which aims to reduce the distance among the distributions of each aspect, will conflict with the sentence reconstruction loss. As a result, it may be hard to obtain a reliable intersection in the attribute space.

Computational resources also have an impact on our approach, as our aspect gap loss leverages a batch-level estimation for each aspect. Therefore, a larger batch size means a more accurate approximation, leaving the attribute space fewer biases. An alternative strategy for smaller batches is to backpropagate the loss after accumulating enough distributional samples, which requires more training epochs.

\section*{Ethics Statement}
We are totally aware that text generation technology has a potential to  be used maliciously to generate fake, toxic, or offensive content. However, after training on the Detoxification aspect, controllable text generation technology is a powerful weapon for combating hate speech, and eliminating harmful information in pre-trained language models. In addition, our multi-aspect controllable text generation technology can take Detoxification as an default aspect when controlling other aspects. We believe it meaningful and beneficial to advance research on controllable text generation.

\section*{Acknowledgements}
Xiaocheng Feng is the corresponding author of this work. We thank the anonymous reviewers for their insightful comments. This work was supported by the National Key R\&D Program of China via grant 2020AAA0106502, National Natural Science Foundation of China (NSFC) via grant 62276078 and the Major Key Project of PCL, PCL2021A06.

\normalem
\bibliography{anthology,custom}

\begin{thebibliography}{36}
\expandafter\ifx\csname natexlab\endcsname\relax\def\natexlab#1{#1}\fi

\bibitem[{Brown et~al.(2020{\natexlab{a}})Brown, Mann, Ryder, Subbiah, Kaplan,
  Dhariwal, Neelakantan, Shyam, Sastry, Askell, Agarwal, Herbert-Voss, Krueger,
  Henighan, Child, Ramesh, Ziegler, Wu, Winter, Hesse, Chen, Sigler, Litwin,
  Gray, Chess, Clark, Berner, McCandlish, Radford, Sutskever, and
  Amodei}]{NEURIPS2020_1457c0d6}
Tom Brown, Benjamin Mann, Nick Ryder, Melanie Subbiah, Jared~D Kaplan, Prafulla
  Dhariwal, Arvind Neelakantan, Pranav Shyam, Girish Sastry, Amanda Askell,
  Sandhini Agarwal, Ariel Herbert-Voss, Gretchen Krueger, Tom Henighan, Rewon
  Child, Aditya Ramesh, Daniel Ziegler, Jeffrey Wu, Clemens Winter, Chris
  Hesse, Mark Chen, Eric Sigler, Mateusz Litwin, Scott Gray, Benjamin Chess,
  Jack Clark, Christopher Berner, Sam McCandlish, Alec Radford, Ilya Sutskever,
  and Dario Amodei. 2020{\natexlab{a}}.
\newblock \href
  {https://proceedings.neurips.cc/paper/2020/file/1457c0d6bfcb4967418bfb8ac142f64a-Paper.pdf}
  {Language models are few-shot learners}.
\newblock In \emph{Advances in Neural Information Processing Systems},
  volume~33, pages 1877--1901. Curran Associates, Inc.

\bibitem[{Brown et~al.(2020{\natexlab{b}})Brown, Mann, Ryder, Subbiah, Kaplan,
  Dhariwal, Neelakantan, Shyam, Sastry, Askell et~al.}]{brown2020language}
Tom Brown, Benjamin Mann, Nick Ryder, Melanie Subbiah, Jared~D Kaplan, Prafulla
  Dhariwal, Arvind Neelakantan, Pranav Shyam, Girish Sastry, Amanda Askell,
  et~al. 2020{\natexlab{b}}.
\newblock Language models are few-shot learners.
\newblock \emph{Advances in neural information processing systems},
  33:1877--1901.

\bibitem[{Carlsson et~al.(2022)Carlsson, {\"O}hman, Liu, Verlinden, Nivre, and
  Sahlgren}]{carlsson-etal-2022-fine}
Fredrik Carlsson, Joey {\"O}hman, Fangyu Liu, Severine Verlinden, Joakim Nivre,
  and Magnus Sahlgren. 2022.
\newblock \href {https://aclanthology.org/2022.acl-long.471} {Fine-grained
  controllable text generation using non-residual prompting}.
\newblock In \emph{Proceedings of the 60th Annual Meeting of the Association
  for Computational Linguistics (Volume 1: Long Papers)}, pages 6837--6857,
  Dublin, Ireland. Association for Computational Linguistics.

\bibitem[{Dathathri et~al.(2020)Dathathri, Madotto, Lan, Hung, Frank, Molino,
  Yosinski, and Liu}]{Dathathri2020Plug}
Sumanth Dathathri, Andrea Madotto, Janice Lan, Jane Hung, Eric Frank, Piero
  Molino, Jason Yosinski, and Rosanne Liu. 2020.
\newblock \href {https://openreview.net/forum?id=H1edEyBKDS} {Plug and play
  language models: A simple approach to controlled text generation}.
\newblock In \emph{International Conference on Learning Representations}.

\bibitem[{Duan et~al.(2020)Duan, Xu, Pei, Han, and Li}]{duan-etal-2020-pre}
Yu~Duan, Canwen Xu, Jiaxin Pei, Jialong Han, and Chenliang Li. 2020.
\newblock \href {https://doi.org/10.18653/v1/2020.acl-main.23} {Pre-train and
  plug-in: Flexible conditional text generation with variational
  auto-encoders}.
\newblock In \emph{Proceedings of the 58th Annual Meeting of the Association
  for Computational Linguistics}, pages 253--262, Online. Association for
  Computational Linguistics.

\bibitem[{Ficler and Goldberg(2017)}]{ficler-goldberg-2017-controlling}
Jessica Ficler and Yoav Goldberg. 2017.
\newblock \href {https://doi.org/10.18653/v1/W17-4912} {Controlling linguistic
  style aspects in neural language generation}.
\newblock In \emph{Proceedings of the Workshop on Stylistic Variation}, pages
  94--104, Copenhagen, Denmark. Association for Computational Linguistics.

\bibitem[{Gu et~al.(2022)Gu, Feng, Ma, Wu, Gong, and
  Qin}]{gu-etal-2022-improving}
Yuxuan Gu, Xiaocheng Feng, Sicheng Ma, Jiaming Wu, Heng Gong, and Bing Qin.
  2022.
\newblock \href {https://aclanthology.org/2022.findings-acl.272} {Improving
  controllable text generation with position-aware weighted decoding}.
\newblock In \emph{Findings of the Association for Computational Linguistics:
  ACL 2022}, pages 3449--3467, Dublin, Ireland. Association for Computational
  Linguistics.

\bibitem[{He et~al.(2021{\natexlab{a}})He, Gao, and Chen}]{he2021debertav3}
Pengcheng He, Jianfeng Gao, and Weizhu Chen. 2021{\natexlab{a}}.
\newblock \href {https://arxiv.org/abs/2111.09543} {Debertav3: Improving
  deberta using electra-style pre-training with gradient-disentangled embedding
  sharing}.
\newblock \emph{arXiv preprint arXiv:2111.09543}.

\bibitem[{He et~al.(2021{\natexlab{b}})He, Liu, Gao, and Chen}]{he2021deberta}
Pengcheng He, Xiaodong Liu, Jianfeng Gao, and Weizhu Chen. 2021{\natexlab{b}}.
\newblock \href {https://openreview.net/forum?id=XPZIaotutsD} {Deberta:
  Decoding-enhanced bert with disentangled attention}.
\newblock In \emph{International Conference on Learning Representations}.

\bibitem[{Hu et~al.(2017)Hu, Yang, Liang, Salakhutdinov, and
  Xing}]{10.5555/3305381.3305545}
Zhiting Hu, Zichao Yang, Xiaodan Liang, Ruslan Salakhutdinov, and Eric~P. Xing.
  2017.
\newblock Toward controlled generation of text.
\newblock In \emph{Proceedings of the 34th International Conference on Machine
  Learning - Volume 70}, ICML'17, page 1587–1596. JMLR.org.

\bibitem[{Keskar et~al.(2019)Keskar, McCann, Varshney, Xiong, and
  Socher}]{keskarCTRL2019}
Nitish~Shirish Keskar, Bryan McCann, Lav Varshney, Caiming Xiong, and Richard
  Socher. 2019.
\newblock {CTRL - A Conditional Transformer Language Model for Controllable
  Generation}.
\newblock \emph{arXiv preprint arXiv:1909.05858}.

\bibitem[{Khalifa et~al.(2020)Khalifa, Elsahar, and
  Dymetman}]{khalifa2020distributional}
Muhammad Khalifa, Hady Elsahar, and Marc Dymetman. 2020.
\newblock A distributional approach to controlled text generation.
\newblock In \emph{International Conference on Learning Representations}.

\bibitem[{Krause et~al.(2021)Krause, Gotmare, McCann, Keskar, Joty, Socher, and
  Rajani}]{krause-etal-2021-gedi-generative}
Ben Krause, Akhilesh~Deepak Gotmare, Bryan McCann, Nitish~Shirish Keskar,
  Shafiq Joty, Richard Socher, and Nazneen~Fatema Rajani. 2021.
\newblock \href {https://doi.org/10.18653/v1/2021.findings-emnlp.424}
  {{G}e{D}i: Generative discriminator guided sequence generation}.
\newblock In \emph{Findings of the Association for Computational Linguistics:
  EMNLP 2021}, pages 4929--4952, Punta Cana, Dominican Republic. Association
  for Computational Linguistics.

\bibitem[{Kumar et~al.(2021)Kumar, Malmi, Severyn, and
  Tsvetkov}]{kumar2021controlled}
Sachin Kumar, Eric Malmi, Aliaksei Severyn, and Yulia Tsvetkov. 2021.
\newblock Controlled text generation as continuous optimization with multiple
  constraints.
\newblock \emph{Advances in Neural Information Processing Systems}, 34.

\bibitem[{Kumar et~al.(2022)Kumar, Paria, and Tsvetkov}]{kumar2022constrained}
Sachin Kumar, Biswajit Paria, and Yulia Tsvetkov. 2022.
\newblock Constrained sampling from language models via langevin dynamics in
  embedding spaces.
\newblock \emph{arXiv preprint arXiv:2205.12558}.

\bibitem[{LeCun et~al.(2006)LeCun, Chopra, Hadsell, Ranzato, and
  Huang}]{lecun2006tutorial}
Yann LeCun, Sumit Chopra, Raia Hadsell, M~Ranzato, and F~Huang. 2006.
\newblock A tutorial on energy-based learning.

\bibitem[{Li et~al.(2016)Li, Galley, Brockett, Gao, and
  Dolan}]{li-etal-2016-diversity}
Jiwei Li, Michel Galley, Chris Brockett, Jianfeng Gao, and Bill Dolan. 2016.
\newblock \href {https://doi.org/10.18653/v1/N16-1014} {A diversity-promoting
  objective function for neural conversation models}.
\newblock In \emph{Proceedings of the 2016 Conference of the North {A}merican
  Chapter of the Association for Computational Linguistics: Human Language
  Technologies}, pages 110--119, San Diego, California. Association for
  Computational Linguistics.

\bibitem[{Li and Liang(2021)}]{li-liang-2021-prefix}
Xiang~Lisa Li and Percy Liang. 2021.
\newblock \href {https://doi.org/10.18653/v1/2021.acl-long.353} {Prefix-tuning:
  Optimizing continuous prompts for generation}.
\newblock In \emph{Proceedings of the 59th Annual Meeting of the Association
  for Computational Linguistics and the 11th International Joint Conference on
  Natural Language Processing (Volume 1: Long Papers)}, pages 4582--4597,
  Online. Association for Computational Linguistics.

\bibitem[{Lin and Riedl(2021)}]{lin-riedl-2021-plug}
Zhiyu Lin and Mark Riedl. 2021.
\newblock \href {https://doi.org/10.18653/v1/2021.nuse-1.7} {Plug-and-blend: A
  framework for controllable story generation with blended control codes}.
\newblock In \emph{Proceedings of the Third Workshop on Narrative
  Understanding}, pages 62--71, Virtual. Association for Computational
  Linguistics.

\bibitem[{Liu et~al.(2021{\natexlab{a}})Liu, Sap, Lu, Swayamdipta, Bhagavatula,
  Smith, and Choi}]{liu-etal-2021-dexperts}
Alisa Liu, Maarten Sap, Ximing Lu, Swabha Swayamdipta, Chandra Bhagavatula,
  Noah~A. Smith, and Yejin Choi. 2021{\natexlab{a}}.
\newblock \href {https://doi.org/10.18653/v1/2021.acl-long.522} {{DE}xperts:
  Decoding-time controlled text generation with experts and anti-experts}.
\newblock In \emph{Proceedings of the 59th Annual Meeting of the Association
  for Computational Linguistics and the 11th International Joint Conference on
  Natural Language Processing (Volume 1: Long Papers)}, pages 6691--6706,
  Online. Association for Computational Linguistics.

\bibitem[{Liu et~al.(2021{\natexlab{b}})Liu, Yuan, Fu, Jiang, Hayashi, and
  Neubig}]{liu2021pre}
Pengfei Liu, Weizhe Yuan, Jinlan Fu, Zhengbao Jiang, Hiroaki Hayashi, and
  Graham Neubig. 2021{\natexlab{b}}.
\newblock \href {https://arxiv.org/abs/2107.13586} {Pre-train, prompt, and
  predict: A systematic survey of prompting methods in natural language
  processing}.
\newblock \emph{arXiv preprint arXiv:2107.13586}.

\bibitem[{Maas et~al.(2011)Maas, Daly, Pham, Huang, Ng, and
  Potts}]{maas-etal-2011-learning}
Andrew~L. Maas, Raymond~E. Daly, Peter~T. Pham, Dan Huang, Andrew~Y. Ng, and
  Christopher Potts. 2011.
\newblock \href {https://aclanthology.org/P11-1015} {Learning word vectors for
  sentiment analysis}.
\newblock In \emph{Proceedings of the 49th Annual Meeting of the Association
  for Computational Linguistics: Human Language Technologies}, pages 142--150,
  Portland, Oregon, USA. Association for Computational Linguistics.

\bibitem[{Mai et~al.(2020)Mai, Pappas, Montero, Smith, and
  Henderson}]{mai-etal-2020-plug}
Florian Mai, Nikolaos Pappas, Ivan Montero, Noah~A. Smith, and James Henderson.
  2020.
\newblock \href {https://doi.org/10.18653/v1/2020.emnlp-main.491} {Plug and
  play autoencoders for conditional text generation}.
\newblock In \emph{Proceedings of the 2020 Conference on Empirical Methods in
  Natural Language Processing (EMNLP)}, pages 6076--6092, Online. Association
  for Computational Linguistics.

\bibitem[{Mireshghallah et~al.(2022)Mireshghallah, Goyal, and
  Berg-Kirkpatrick}]{mireshghallah-etal-2022-mix}
Fatemehsadat Mireshghallah, Kartik Goyal, and Taylor Berg-Kirkpatrick. 2022.
\newblock \href {https://aclanthology.org/2022.acl-long.31} {Mix and match:
  Learning-free controllable text generationusing energy language models}.
\newblock In \emph{Proceedings of the 60th Annual Meeting of the Association
  for Computational Linguistics (Volume 1: Long Papers)}, pages 401--415,
  Dublin, Ireland. Association for Computational Linguistics.

\bibitem[{Pan et~al.(2010)Pan, Ni, Sun, Yang, and
  Chen}]{10.1145/1772690.1772767}
Sinno~Jialin Pan, Xiaochuan Ni, Jian-Tao Sun, Qiang Yang, and Zheng Chen. 2010.
\newblock \href {https://doi.org/10.1145/1772690.1772767} {Cross-domain
  sentiment classification via spectral feature alignment}.
\newblock In \emph{Proceedings of the 19th International Conference on World
  Wide Web}, WWW '10, page 751–760, New York, NY, USA. Association for
  Computing Machinery.

\bibitem[{Pillutla et~al.(2021)Pillutla, Swayamdipta, Zellers, Thickstun,
  Welleck, Choi, and Harchaoui}]{NEURIPS2021_260c2432}
Krishna Pillutla, Swabha Swayamdipta, Rowan Zellers, John Thickstun, Sean
  Welleck, Yejin Choi, and Zaid Harchaoui. 2021.
\newblock \href
  {https://proceedings.neurips.cc/paper/2021/file/260c2432a0eecc28ce03c10dadc078a4-Paper.pdf}
  {Mauve: Measuring the gap between neural text and human text using divergence
  frontiers}.
\newblock In \emph{Advances in Neural Information Processing Systems},
  volume~34, pages 4816--4828. Curran Associates, Inc.

\bibitem[{Qian et~al.(2022)Qian, Dong, Shen, Wei, and
  Chen}]{qian-etal-2022-controllable}
Jing Qian, Li~Dong, Yelong Shen, Furu Wei, and Weizhu Chen. 2022.
\newblock \href {https://aclanthology.org/2022.findings-acl.229} {Controllable
  natural language generation with contrastive prefixes}.
\newblock In \emph{Findings of the Association for Computational Linguistics:
  ACL 2022}, pages 2912--2924, Dublin, Ireland. Association for Computational
  Linguistics.

\bibitem[{Qin et~al.(2022)Qin, Welleck, Khashabi, and Choi}]{qin2022cold}
Lianhui Qin, Sean Welleck, Daniel Khashabi, and Yejin Choi. 2022.
\newblock Cold decoding: Energy-based constrained text generation with langevin
  dynamics.
\newblock \emph{arXiv preprint arXiv:2202.11705}.

\bibitem[{Radford et~al.(2019)Radford, Wu, Child, Luan, Amodei, and
  Sutskever}]{radford2019language}
Alec Radford, Jeff Wu, Rewon Child, David Luan, Dario Amodei, and Ilya
  Sutskever. 2019.
\newblock Language models are unsupervised multitask learners.

\bibitem[{Wu et~al.(2022)Wu, Motamed, Srivastava, and De~la
  Torre}]{wu2022generative}
Chen~Henry Wu, Saman Motamed, Shaunak Srivastava, and Fernando De~la Torre.
  2022.
\newblock Generative visual prompt: Unifying distributional control of
  pre-trained generative models.
\newblock \emph{arXiv preprint arXiv:2209.06970}.

\bibitem[{Yang and Klein(2021)}]{yang-klein-2021-fudge}
Kevin Yang and Dan Klein. 2021.
\newblock \href {https://doi.org/10.18653/v1/2021.naacl-main.276} {{FUDGE}:
  Controlled text generation with future discriminators}.
\newblock In \emph{Proceedings of the 2021 Conference of the North American
  Chapter of the Association for Computational Linguistics: Human Language
  Technologies}, pages 3511--3535, Online. Association for Computational
  Linguistics.

\bibitem[{Yang et~al.(2022)Yang, Liu, Lei, Yang, Xue, Chen, and
  Xie}]{yang2022tailor}
Kexin Yang, Dayiheng Liu, Wenqiang Lei, Baosong Yang, Mingfeng Xue, Boxing
  Chen, and Jun Xie. 2022.
\newblock Tailor: A prompt-based approach to attribute-based controlled text
  generation.
\newblock \emph{arXiv preprint arXiv:2204.13362}.

\bibitem[{Yu et~al.(2021)Yu, Yu, and Sagae}]{yu-etal-2021-attribute-alignment}
Dian Yu, Zhou Yu, and Kenji Sagae. 2021.
\newblock \href {https://doi.org/10.18653/v1/2021.findings-emnlp.194}
  {Attribute alignment: Controlling text generation from pre-trained language
  models}.
\newblock In \emph{Findings of the Association for Computational Linguistics:
  EMNLP 2021}, pages 2251--2268, Punta Cana, Dominican Republic. Association
  for Computational Linguistics.

\bibitem[{Zhang et~al.(2022)Zhang, Roller, Goyal, Artetxe, Chen, Chen, Dewan,
  Diab, Li, Lin et~al.}]{zhang2022opt}
Susan Zhang, Stephen Roller, Naman Goyal, Mikel Artetxe, Moya Chen, Shuohui
  Chen, Christopher Dewan, Mona Diab, Xian Li, Xi~Victoria Lin, et~al. 2022.
\newblock \href {https://arxiv.org/abs/2205.01068} {Opt: Open pre-trained
  transformer language models}.
\newblock \emph{arXiv preprint arXiv:2205.01068}.

\bibitem[{Zhang et~al.(2015)Zhang, Zhao, and LeCun}]{NIPS2015_250cf8b5}
Xiang Zhang, Junbo Zhao, and Yann LeCun. 2015.
\newblock \href
  {https://proceedings.neurips.cc/paper/2015/file/250cf8b51c773f3f8dc8b4be867a9a02-Paper.pdf}
  {Character-level convolutional networks for text classification}.
\newblock In \emph{Advances in Neural Information Processing Systems},
  volume~28. Curran Associates, Inc.

\bibitem[{Ziegler et~al.(2019)Ziegler, Stiennon, Wu, Brown, Radford, Amodei,
  Christiano, and Irving}]{ziegler2019finetuning}
Daniel~M. Ziegler, Nisan Stiennon, Jeffrey Wu, Tom~B. Brown, Alec Radford,
  Dario Amodei, Paul Christiano, and Geoffrey Irving. 2019.
\newblock \href {https://arxiv.org/abs/1909.08593} {Fine-tuning language models
  from human preferences}.
\newblock \emph{arXiv preprint arXiv:1909.08593}.

\end{thebibliography}
\bibliographystyle{acl_natbib}

\clearpage

\appendix

\section{Distribution of Attributes}

\subsection{Independent Projection of Attributes}
\label{sec:appendix1}
We project sample points to 2D by Principal Component Analysis, with each attribute projected independently. We display a scatter plot for each and perform the Gaussian kernel density estimation. The darker area denotes a higher probability, where more representation points of oracle sentences gather. And the region annotated by a red ellipse is the estimated distributional center.

\begin{figure}[h]
  \centering
  \includegraphics[width=\columnwidth]{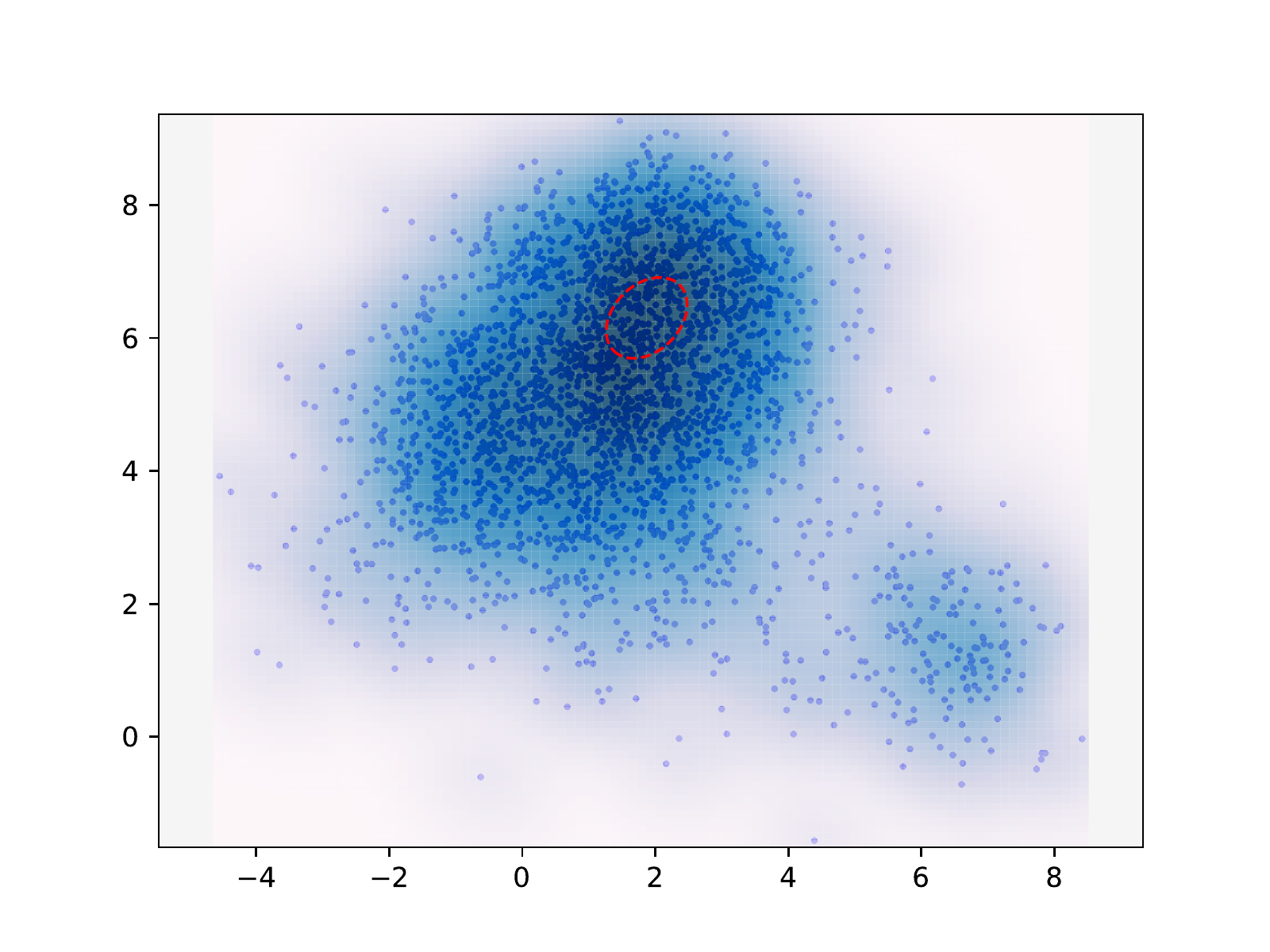}
  \caption{Distribution of World attribute from Topic aspect.}
  \label{appfig:11}
\end{figure}

\begin{figure}[h]
  \centering
  \includegraphics[width=\columnwidth]{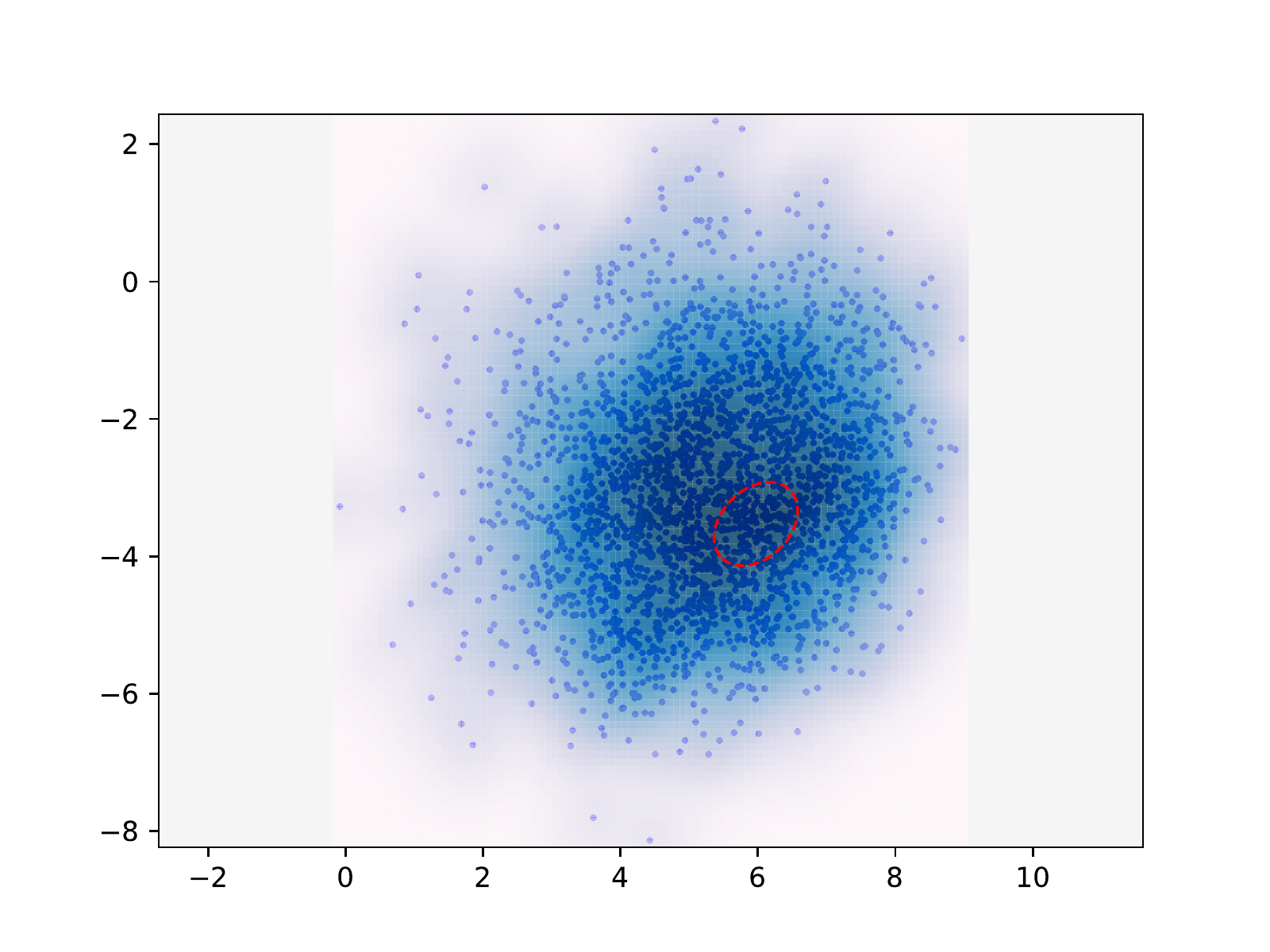}
  \caption{Distribution of Sports attribute from Topic aspect.}
  \label{appfig:12}
\end{figure}

\begin{figure}[h]
  \centering
  \includegraphics[width=\columnwidth]{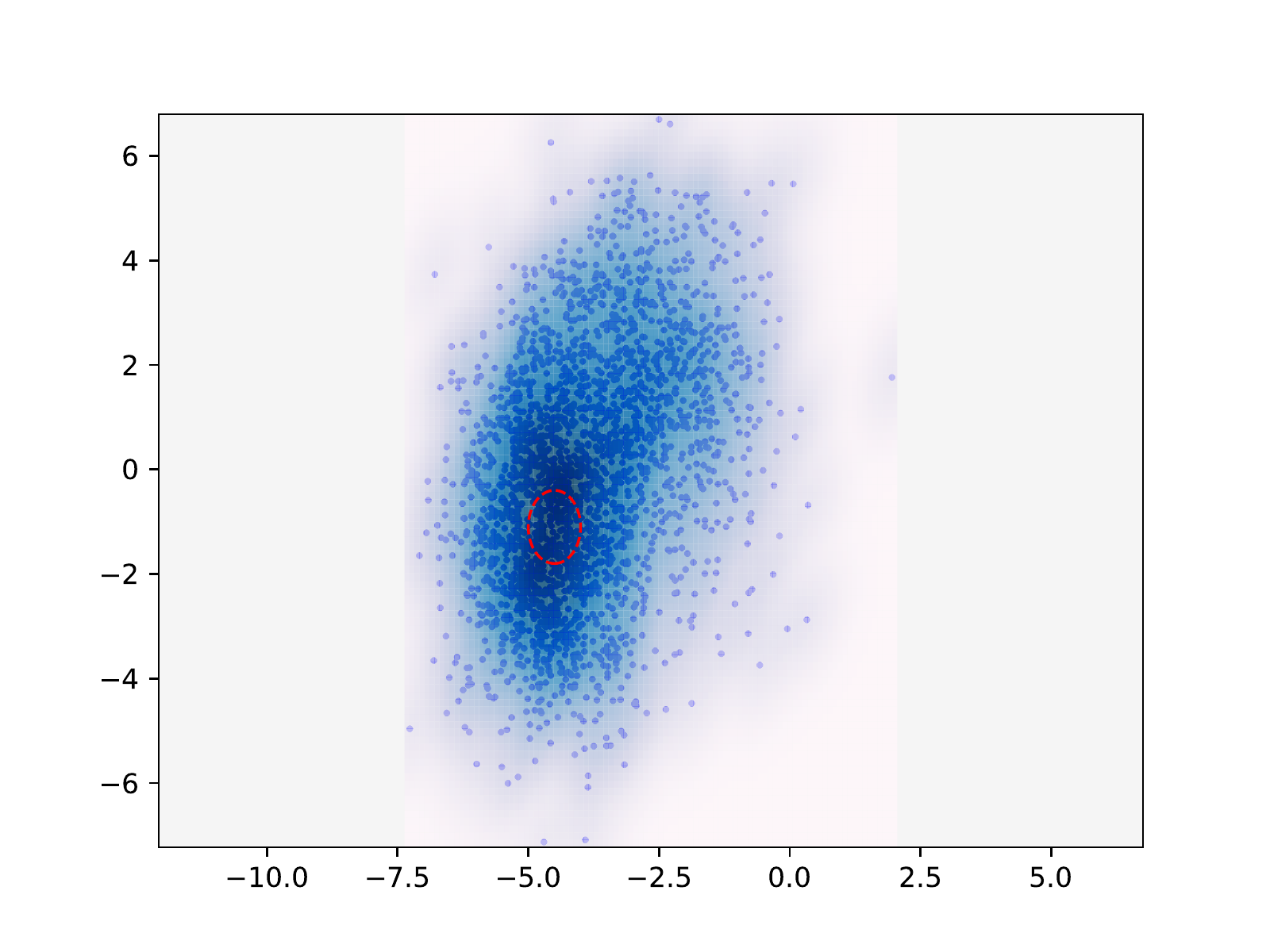}
  \caption{Distribution of Sci/Tech attribute from Topic aspect.}
  \label{appfig:13}
  \vspace{-0.5cm}
\end{figure}

We underline distributions of attributes in \Cref{appfig:11,appfig:12,appfig:13}, including World, Sports, and Sci/Tech, which are significantly asymmetric. And especially, the projected distribution of the World attribute is even non-convex. This observation supports our hypothesis that the practical distributions of attributes are far more complex than symmetric distributions such as Gaussian distribution.

\begin{figure}[h]
  \centering
  \includegraphics[width=\columnwidth]{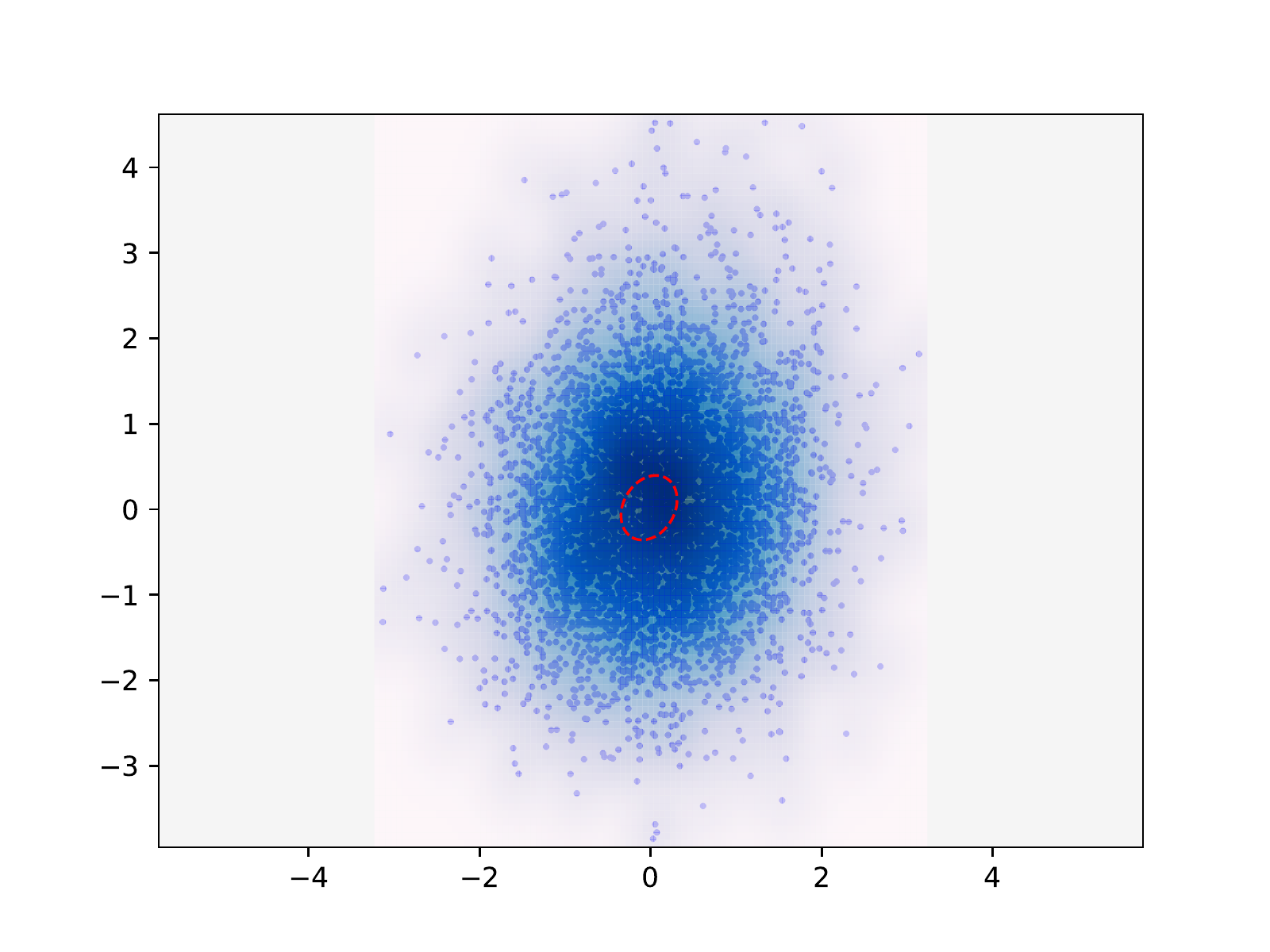}
  \caption{Distribution of Negative attribute from Sentiment aspect.}
  \label{appfig:14}
  \vspace{-0.5cm}
\end{figure}

\begin{figure}[h]
  \centering
  \includegraphics[width=\columnwidth]{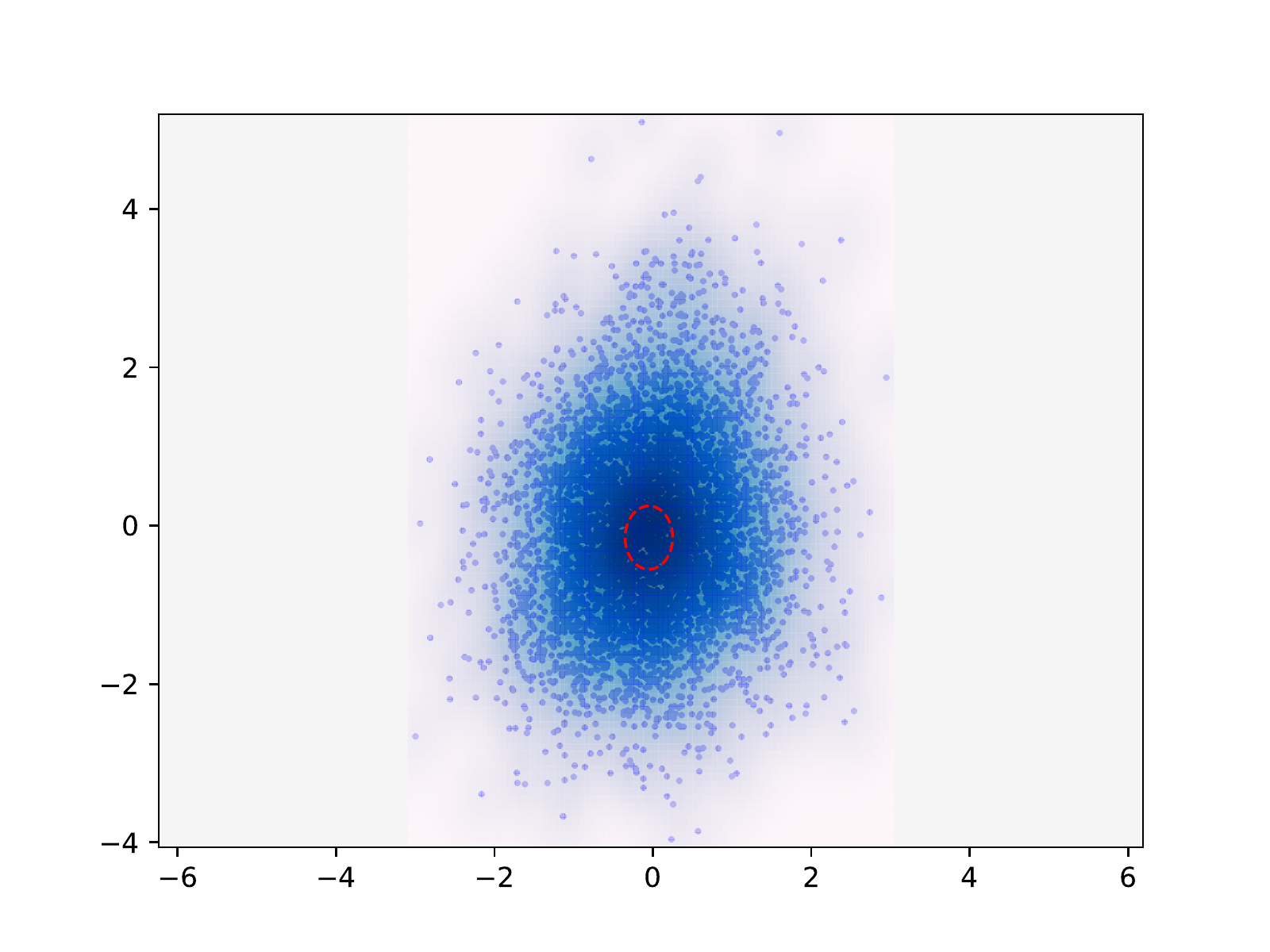}
  \caption{Distribution of Positive attribute from Sentiment aspect.}
  \label{appfig:15}
  \vspace{-0.5cm}
\end{figure}

\begin{figure}[h]
  \centering
  \includegraphics[width=\columnwidth]{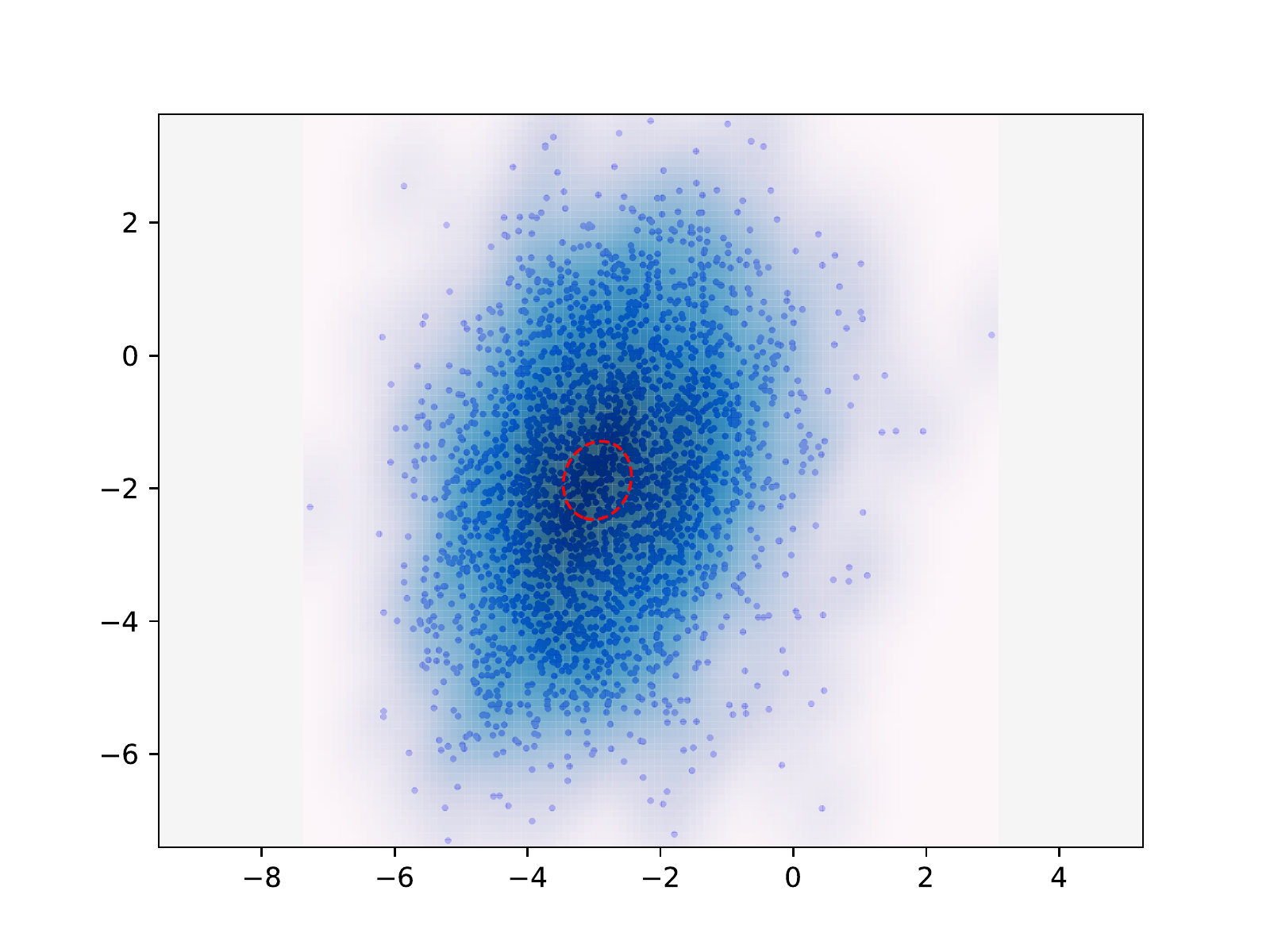}
  \caption{Distribution of Business attribute from Topic aspect.}
  \label{appfig:16}
  \vspace{-0.5cm}
\end{figure}

\begin{figure}[h]
  \centering
  \includegraphics[width=\columnwidth]{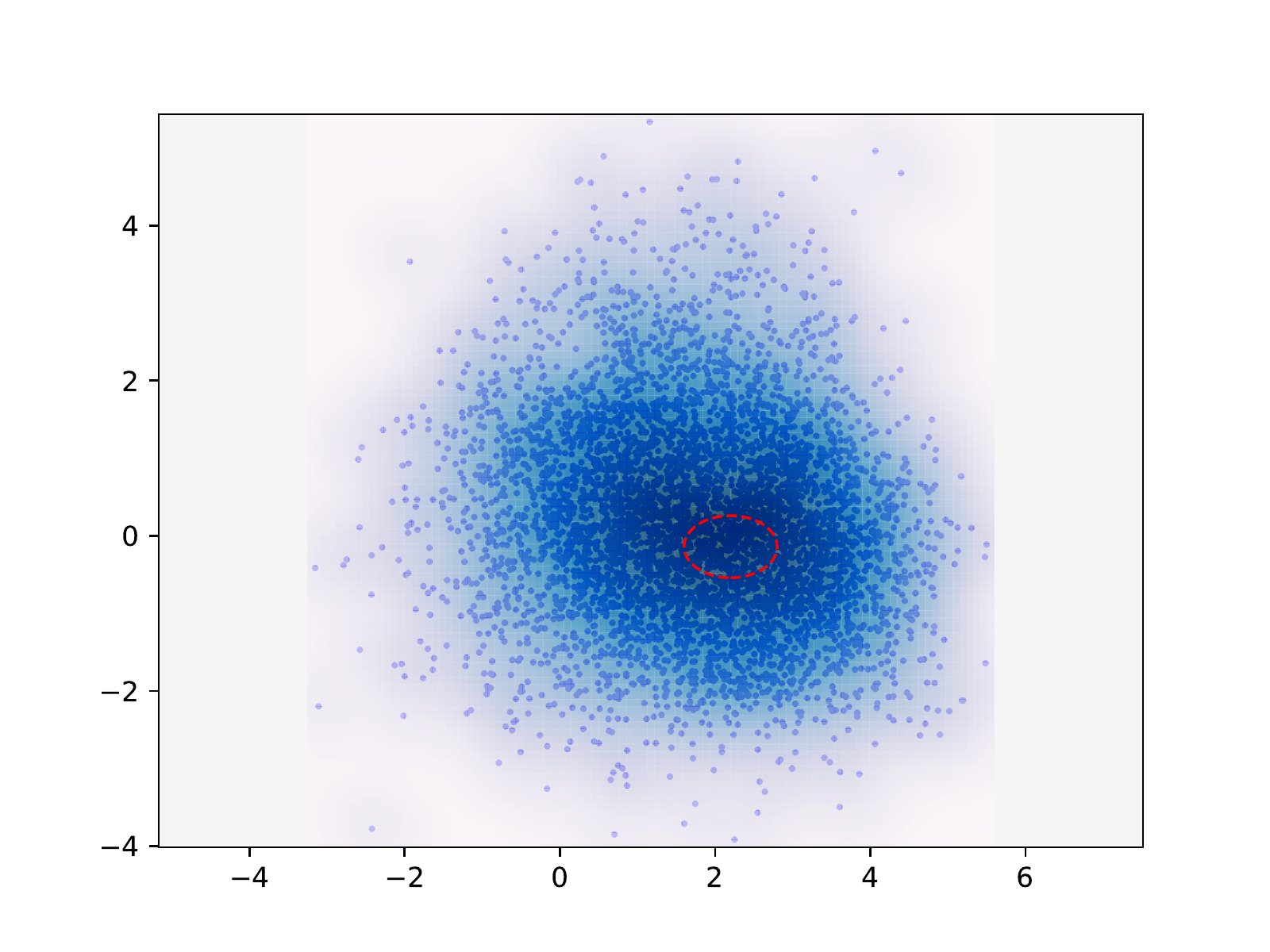}
  \caption{Distribution of Toxic attribute from Detoxification aspect.}
  \label{appfig:17}
  \vspace{-0.5cm}
\end{figure}

\begin{figure}[h]
  \centering
  \includegraphics[width=\columnwidth]{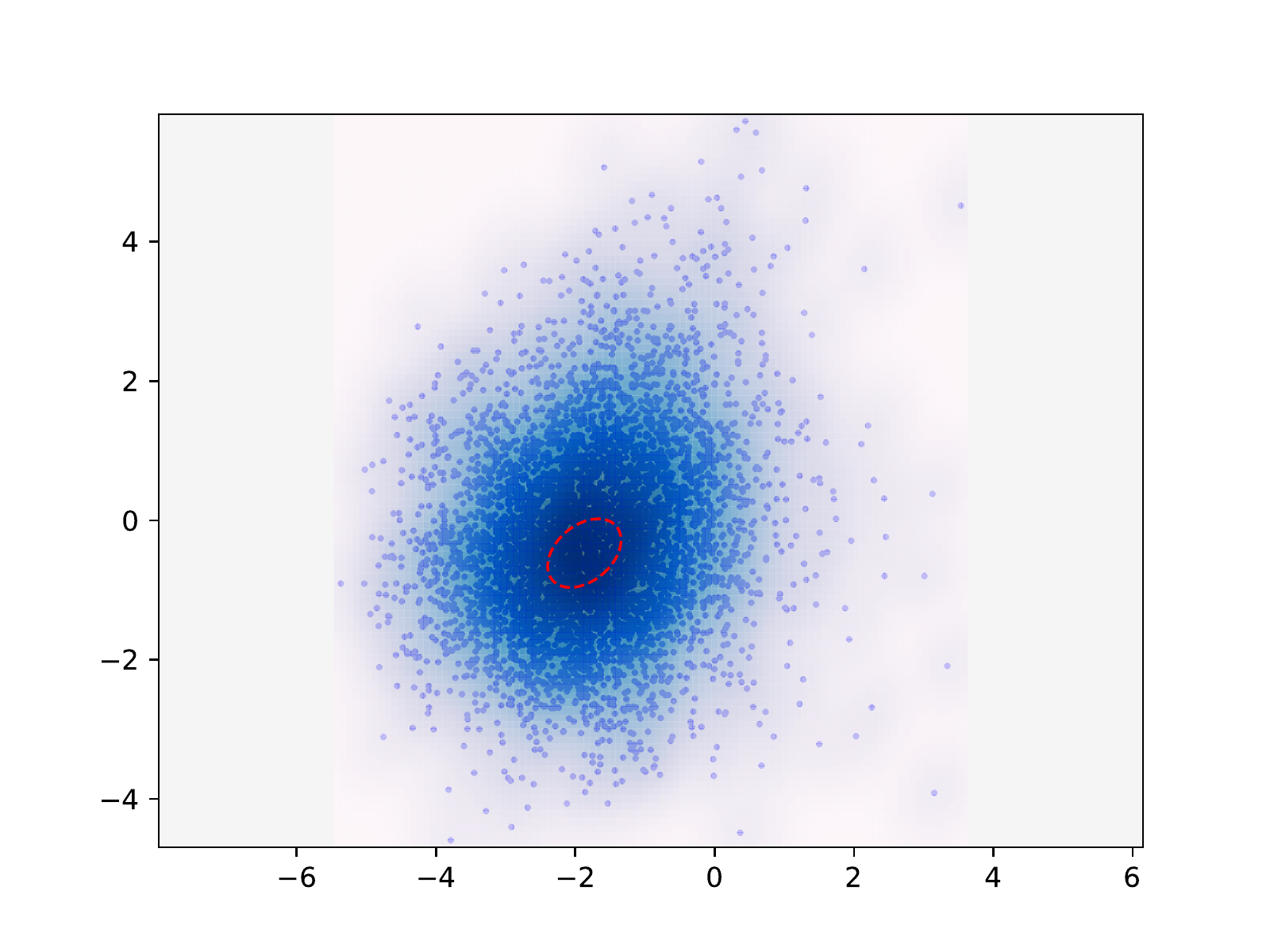}
  \caption{Distribution of Non-toxic attribute from Detoxification aspect.}
  \label{appfig:18}
  \vspace{-0.5cm}
\end{figure}

In addition, we plot projected distributions of other attributes in \Cref{appfig:14,appfig:15,appfig:16,appfig:17,appfig:18}. Attributes such as Positive and Negative seem roughly symmetric in 2D projection. However, we can not guarantee their symmetry in high-dimensional space. Because the PCA aims to identify directions along which the variation in the data is maximal. In other words, the direction selection strategy is not necessarily related to symmetry or asymmetry, which means these 2D symmetric distributions may be asymmetric in high-dimensional space, with the asymmetric directions ignored during projection. Worse still, the long-tail region for a skewed direction may be too sparse, leading to lower variation compared to symmetric directions.

\clearpage
\subsection{Joint Projection of Attributes}
\label{sec:appendix2}

\begin{figure}[h]
  \centering
  \includegraphics[width=\columnwidth]{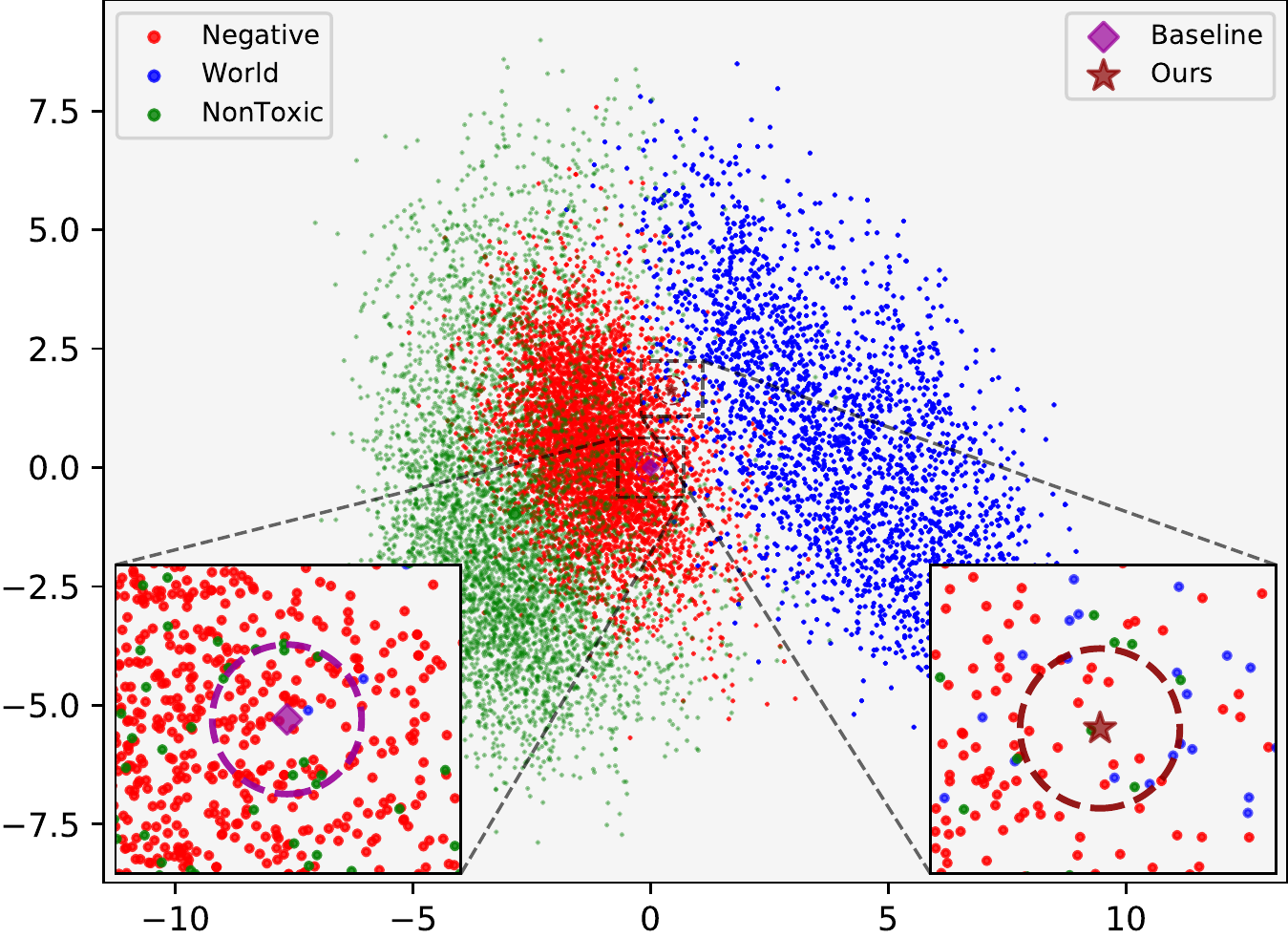}
  \caption{Jointly projected distributions of attributes: Negative, World, and NonToxic from aspects: Sentiment, Topic, and Detoxification, respectively.}
  \label{appfig:21}
\end{figure}

\begin{figure}[h]
  \centering
  \includegraphics[width=\columnwidth]{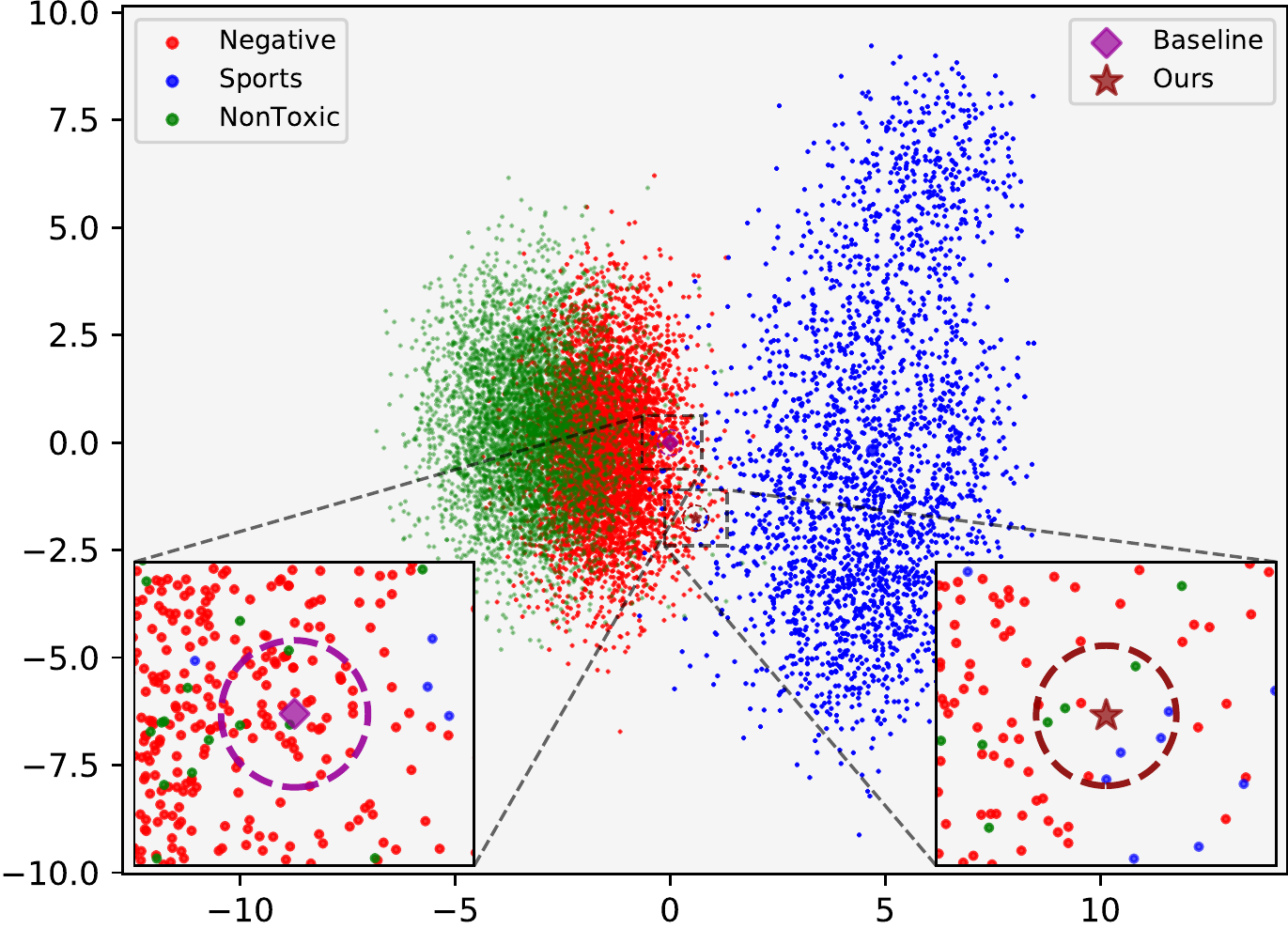}
  \caption{Jointly projected distributions of attributes: Negative, Sports, and NonToxic from aspects: Sentiment, Topic, and Detoxification, respectively.}
  \label{appfig:22}
  \vspace{-0.5cm}
\end{figure}

\begin{figure}[h]
  \centering
  \includegraphics[width=\columnwidth]{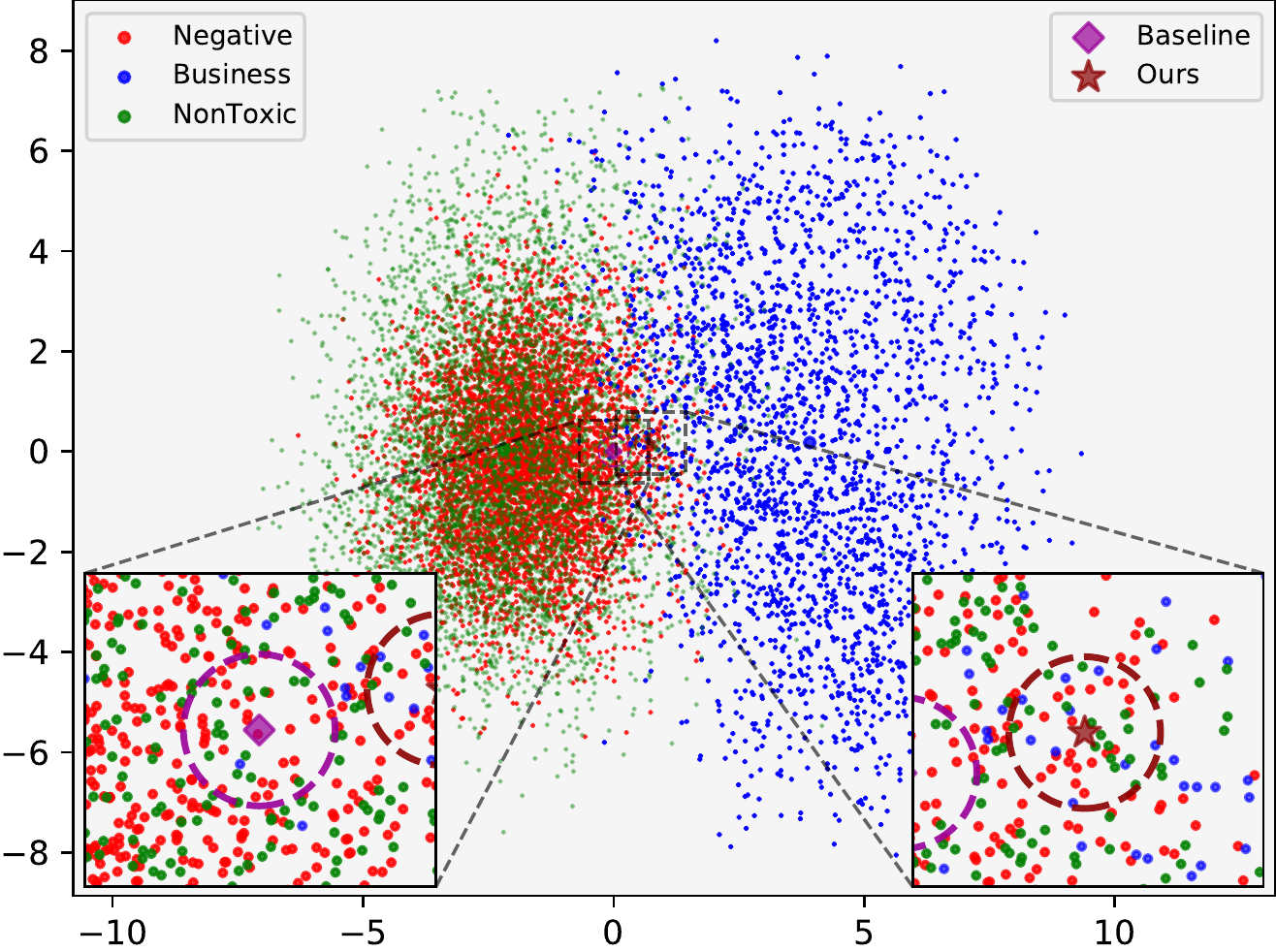}
  \caption{Jointly projected distributions of attributes: Negative, Business, and NonToxic from aspects: Sentiment, Topic, and Detoxification, respectively.}
  \label{appfig:23}
\end{figure}

\begin{figure}[h]
  \centering
  \includegraphics[width=\columnwidth]{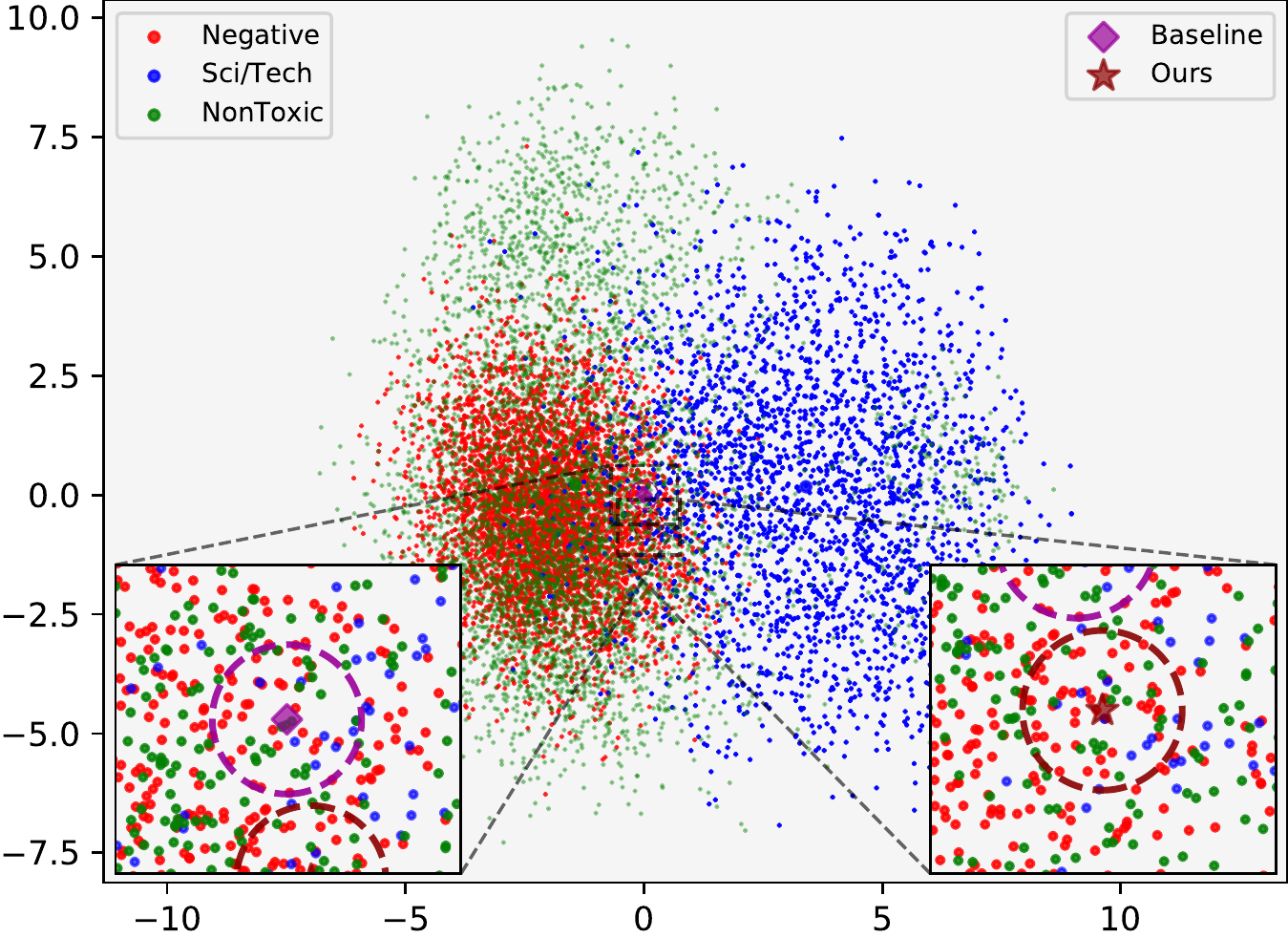}
  \caption{Jointly projected distributions of attributes: Negative, Sci/Tech, and NonToxic from aspects: Sentiment, Topic, and Detoxification, respectively.}
  \label{appfig:24}
\end{figure}

\begin{figure}[h]
  \centering
  \includegraphics[width=\columnwidth]{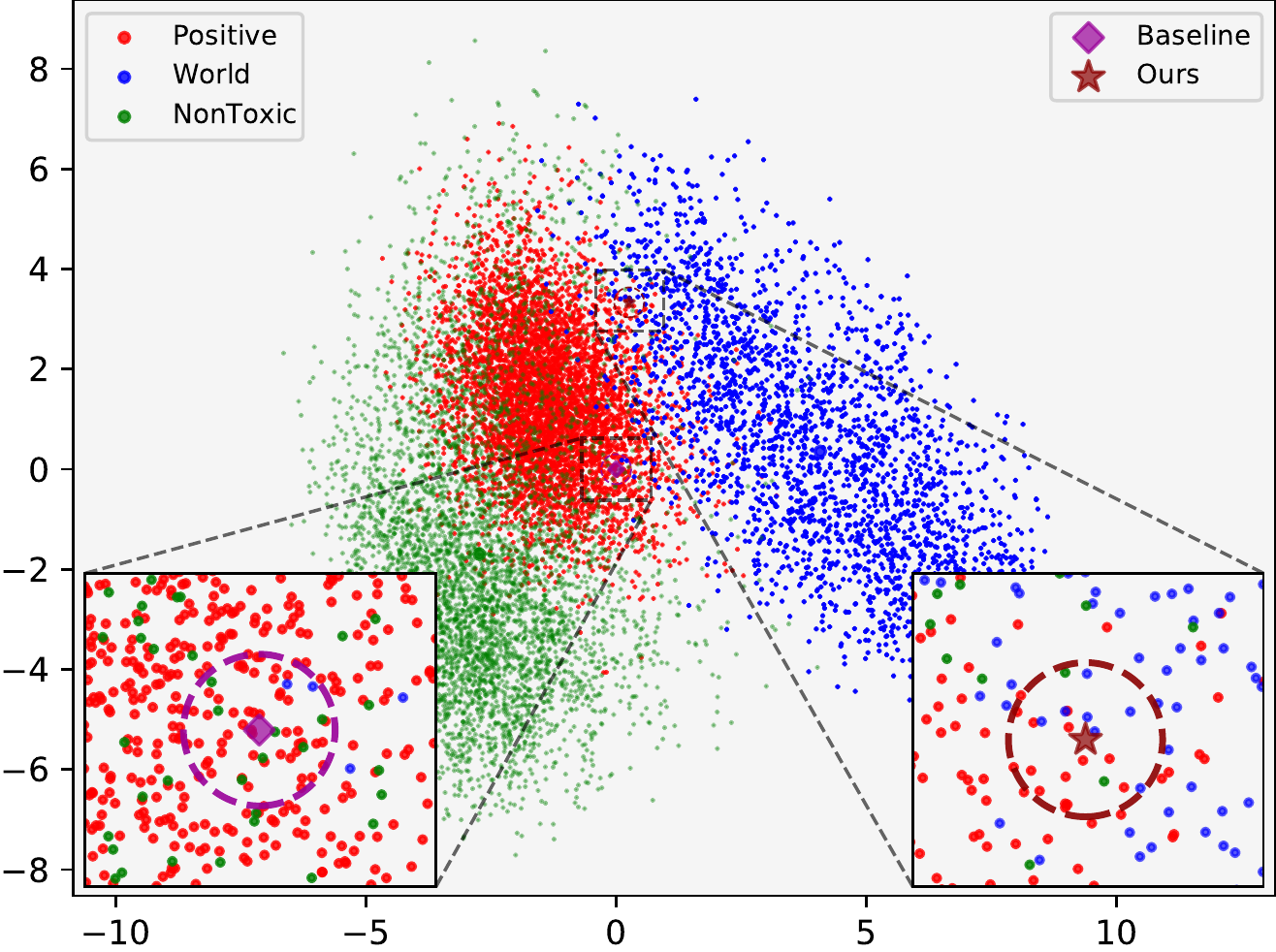}
  \caption{Jointly projected distributions of attributes: Positive, World, and NonToxic from aspects: Sentiment, Topic, and Detoxification, respectively.}
  \label{appfig:25}
\end{figure}

\begin{figure}[H]
  \centering
  \includegraphics[width=\columnwidth]{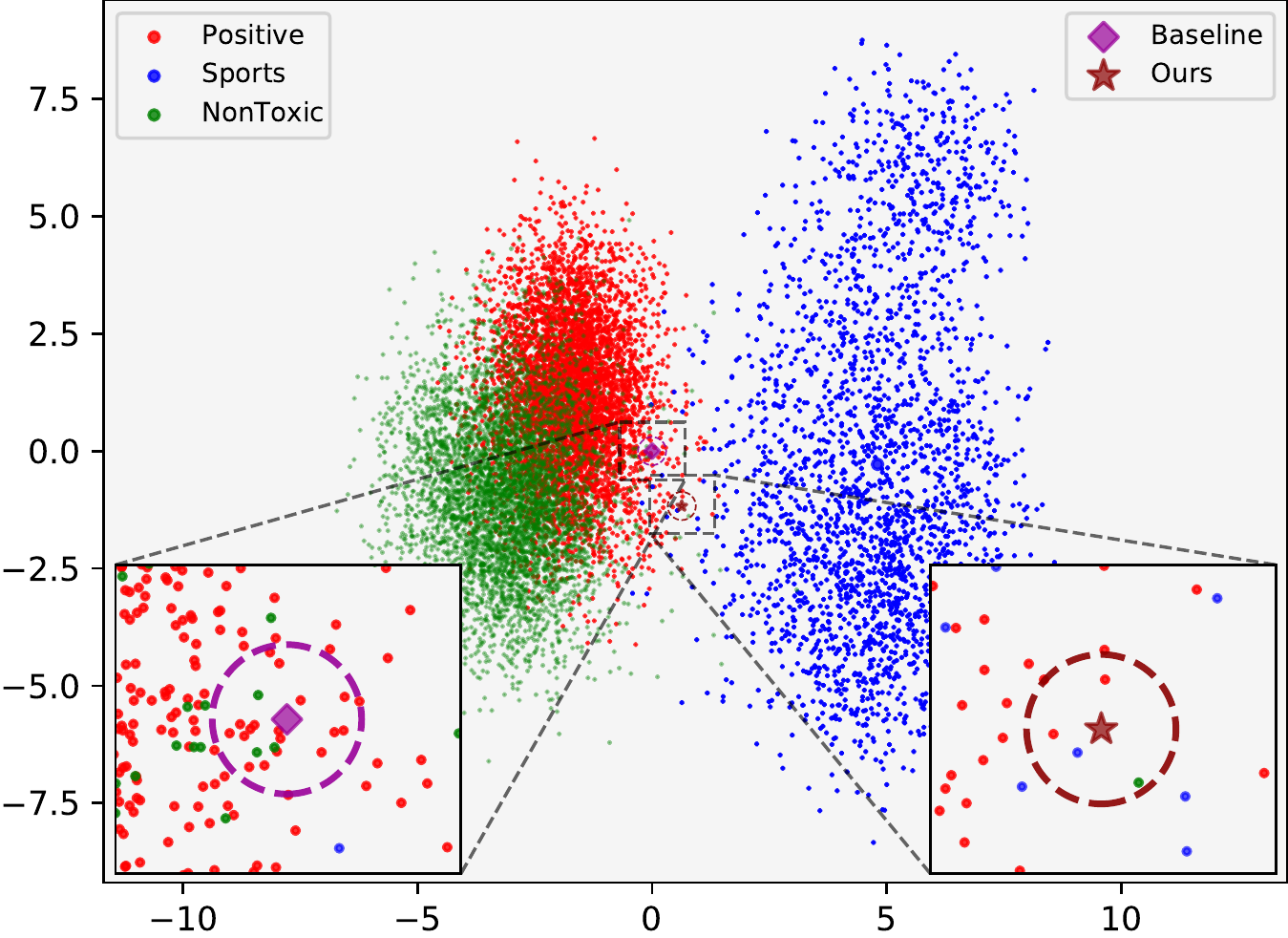}
  \caption{Jointly projected distributions of attributes: Positive, Sports, and NonToxic from aspects: Sentiment, Topic, and Detoxification, respectively.}
  \label{appfig:26}
\end{figure}

\begin{figure}[H]
  \centering
  \includegraphics[width=\columnwidth]{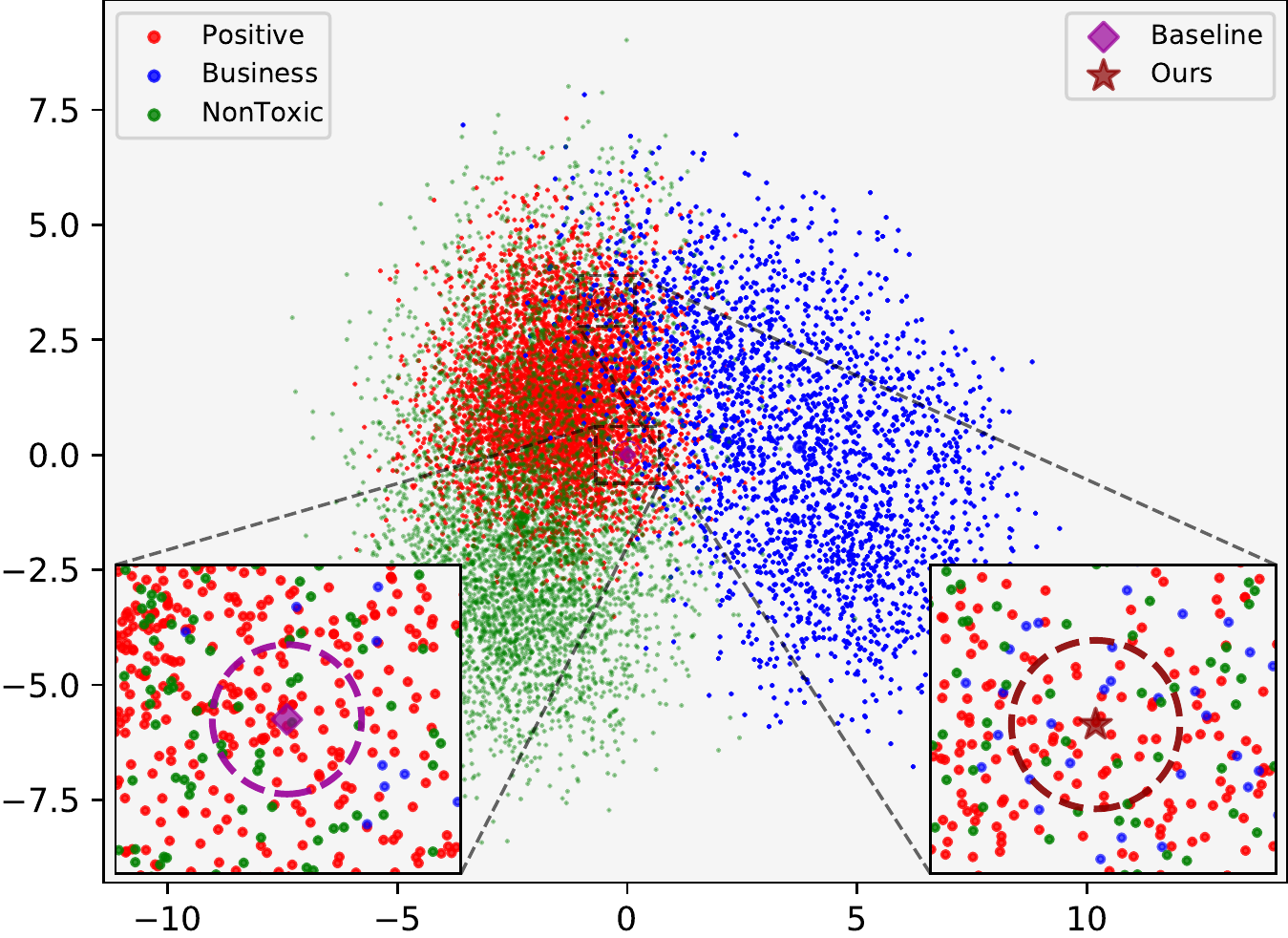}
  \caption{Jointly projected distributions of attributes: Positive, Business, and NonToxic from aspects: Sentiment, Topic, and Detoxification, respectively.}
  \label{appfig:27}
\end{figure}

\begin{figure}[H]
  \centering
  \includegraphics[width=\columnwidth]{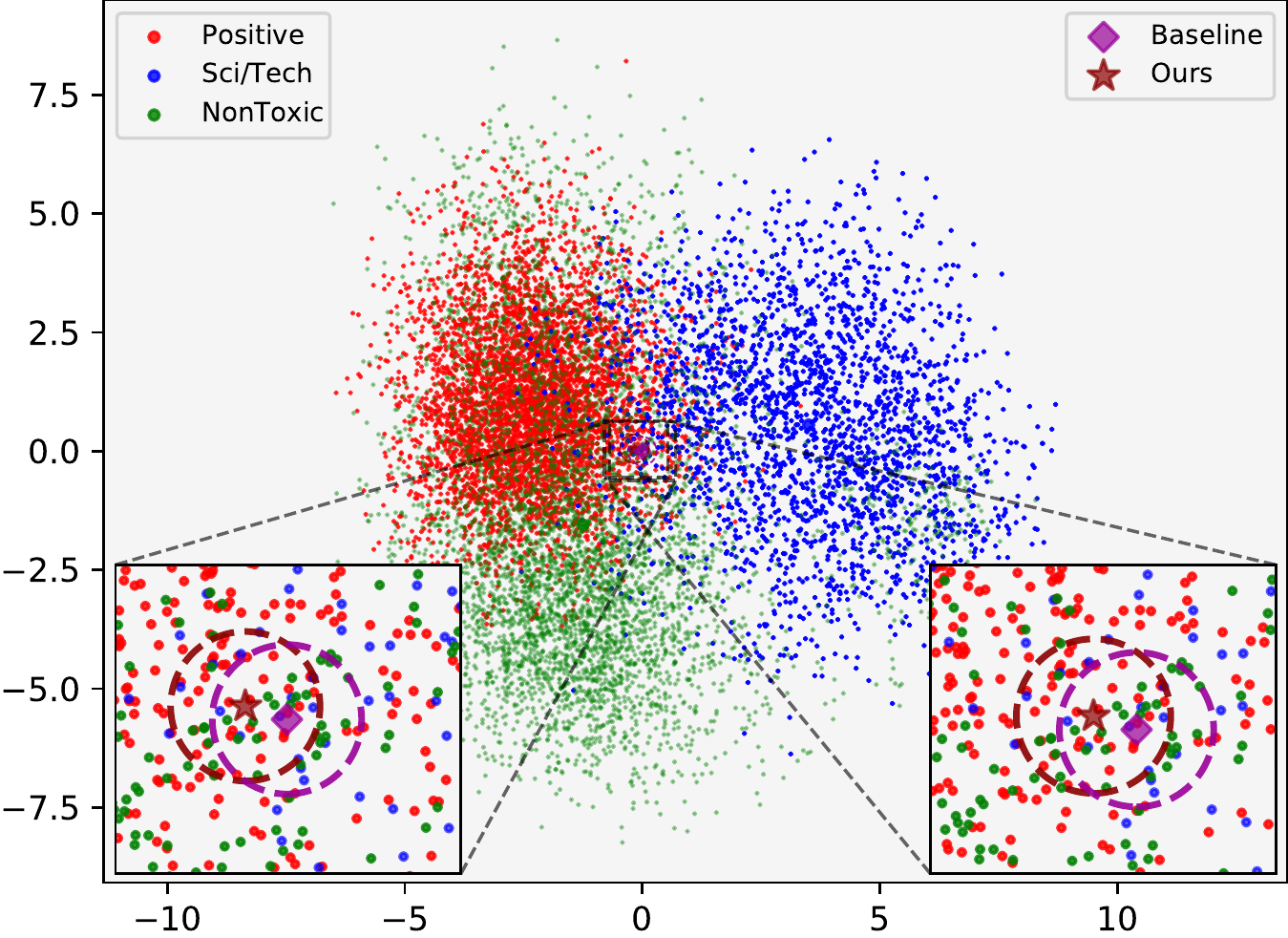}
  \caption{Jointly projected distributions of attributes: Positive, Sci/Tech, and NonToxic from aspects: Sentiment, Topic, and Detoxification, respectively.}
  \label{appfig:28}
\end{figure}

We project combined sample points of attributes from three different aspects jointly to 2D by PCA. We display a scatter plot for each combination in \Cref{appfig:21,appfig:22,appfig:23,appfig:24,appfig:25,appfig:26,appfig:27,appfig:28}.The intersection points calculated on baselines' interpolation strategy and our intersection searching algorithm are plotted with \textbf{\textcolor{Optimal}{Baseline}} and \textbf{\textcolor{darkred}{Ours}}, respectively. From these figures, we observe that NonToxic can mainly cover two sentiment attributes or at least possess large intersection areas. Besides, the intersection areas among sentiment attributes and topic attributes, except for the Sci/Tech, are narrow and sparse. Compared with the baselines' strategy, our search algorithm is closer to the intersection area, especially on Negative and Business attributes in \Cref{appfig:21,appfig:22,appfig:23,appfig:27}.

\clearpage

\newpage
\subsection{Projection of Positive and Negative}
\label{sec:pos_neg}
\begin{figure}[H]
  \centering
  \includegraphics[width=\columnwidth]{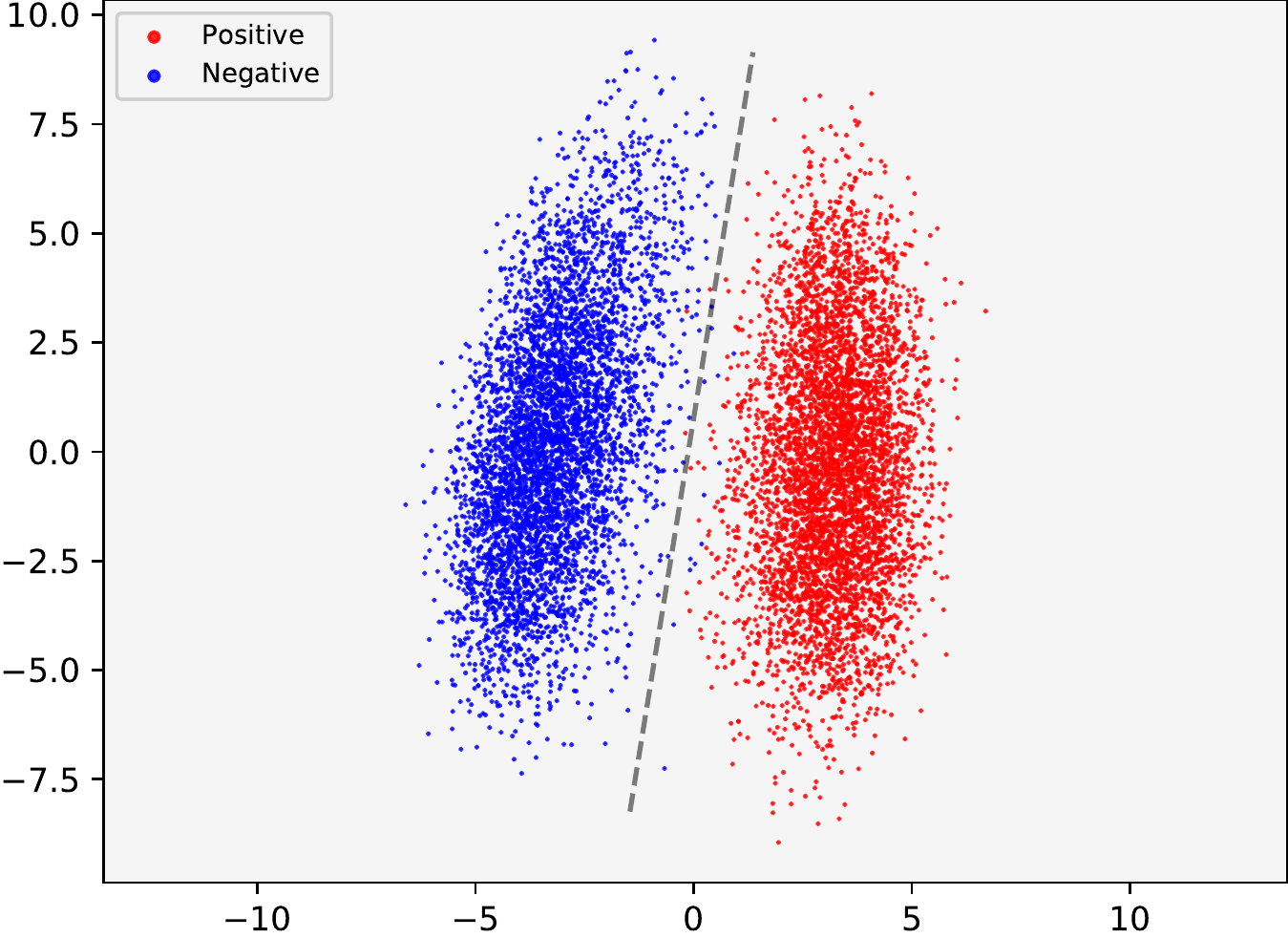}
  \caption{Jointly projected distributions of Positive and Negative.}
  \label{appfig:pos_neg}
\end{figure}
Except for some noise in the dataset, positive and negative do not intersect when jointly projected.

\section{Hyperparameters and Details}
\label{sec:appendix3}

Our methods are implemented using the Hugging face Transformers package. Our encoder is initialized with Bert-base-uncased, and the fixed decoder uses GPT2-medium. For any sentence, it will be tokenized with WordPiece tokenizer from Bert and Byte-Pair Encoding tokenizer from GPT2 before input to encoder and decoder, respectively. We perform mean pooling on outputs of the encoder and convert them to 768-dimensional latent representations, which are points in our attribute space. Afterward, latent representations will be mapped to the prefix with a dimension of $20 \times 24 \times 2 \times 1024$, where $20$ is the prefix sequence length, $24$ is the number of hidden layers in GPT2-medium, $2$ represents one key and one value, and $1024$ is the size of hidden states in GPT2-medium. It's worth noting that prefix length Contrastive Prefix uses for single-aspect control is $10$ and for multi-aspect control is $10 \times \text{number of aspects}$, which is $30$ for three-aspect control. Our prefix length is fixed to $20$, which has nothing to do with the scale of aspects. 

During the training stage, we use half-precision mode for efficiency on one NVIDIA A100 80GB GPU, where the batch size is 128 since the larger batch size better alleviates the aspect gap loss. In our setting, the random seed is $0$, $w_1 = 0.5, w_2 = 0.2, w_3 = 0.3$, variation hyperparameter $\lambda$ is 1e-3, the optimizer is AdamW with a learning rate of 1e-4, the number of training epochs is $150$, and we use a checkpoint at the step $30000$. The training phase takes about 8 hours, and we experiment 6 times to search for the $\lambda \in \left[\text{2e-3, 1e-3, 5e-4, 1e-4, 5e-5, 1e-5}\right]$, while the other hyperparameters are initial settings.

\begin{table}[h]
\small
\centering
\begin{tabular}{l|c}
    \hline
    
    \hline
    \textbf{Combination} & \textbf{Weight}\\
    \hline
    \hline
    Neg. \& World \& NonTox. & $2:7:1$\\
    Neg. \& Sports \& NonTox. & $2:4:1$\\
    Neg. \& Business \& NonTox. & $2:8:1$\\ 
    Neg. \& Sci./Tech. \& NonTox. & $3:1:3$\\
    Pos. \& World \& NonTox. & $2:12:1$\\
    Pos. \& Sports \& NonTox. & $3:5.5:1$\\
    Pos. \& Business \& NonTox. & $2:9:1$\\
    Pos. \& Sci./Tech. \& NonTox. & $3:1:1$\\
    \hline
    
    \hline
\end{tabular}
\caption{Specialized Weight for Attribute Balance.}
\label{apptab:hyper}
\end{table}

During the inference phase, the maximum number of iterations $T$ is $15$, the number of candidates $M$ is $1000$, and the number of neighbors $K$ is $200$. We utilize a specialized list of weight parameters for each combination of attributes in Table \ref{apptab:hyper}, which aims to balance the performance among attributes from different aspects. After the iteration of intersection searching, our strategy is first to select the top 10 candidates with the smallest distances to their neighbors as the final candidate set. Then we randomly choose a candidate from these ten as the intersection's representation for text generation diversity. Our text generation process is the same as prefix tuning with sequence length set to 50. Except for model and data loading, the entire evaluation process for each attribute combination, including intersection searching, text generation, and attribute-relevance evaluation, takes about 2 minutes. Therefore, we can manually tune the weight of attributes to balance them, with a maximum trial number of 8 for each weight.

35 prompts we used in the inferencing stage are following the PPLM setting with 20 from its bag-of-word setting and 15 from its discriminator setting:
\begin{itemize}
\item \textbf{PPLM-Bow}: ``In summary'', ``This essay discusses'', ``Views on'', ``The connection'', ``Foundational to this is'', ``To review,'', ``In brief,'', ``An illustration of'', ``Furthermore,'', ``The central theme'', ``To conclude,'', ``The key aspect'', ``Prior to this'', ``Emphasised are'', ``To summarise'', ``The relationship'', ``More importantly,'', ``It has been shown'', ``The issue focused on'', ``In this essay''.
\item \textbf{PPLM-Discrim}: ``Once upon a time'', ``The book'', ``The chicken'', ``The city'', ``The country'', ``The horse'', ``The lake'', ``The last time'', ``The movie'', ``The painting'', ``The pizza'', ``The potato'', ``The president of the country'', ``The road'', ``The year is 1910.''.
\end{itemize}

Detailed setting of baselines: \begin{enumerate*}[label=(\Roman*)]
\item \textbf{Weighted Decoding}:
For \textbf{PPLM}, we only retrain its classifier heads on our datasets while keeping all other original settings. For \textbf{GeDi}, We use its code directly since we are following its setting. 
\item \textbf{Multi-objective Optimization}:
\textbf{MUCOCO} provides a solution for custom classification constraints, and thus we train these classifiers on our datasets. \textbf{Mix\&Match} is relatively complex as it can not generate long sentences from scratch with the mask language model Bert. Worse still, as a method based on sampling, it is somewhat dependent on initialization. Therefore, we use sentences generated by PPLM as the starting sentences and let Mix\&Match slowly polish the text by itself in iterations. 
\item \textbf{Prefix-Tuning}:
We reproduce \textbf{Contrastive Prefix}\footnote{They didn't make the code public. We reproduce the code and get comparable results to what they report.} and achieve comparable results.
\end{enumerate*}
For a fair comparison, we unify the pre-trained language model to GPT2-medium (345M parameters) except for Mix\&Match using Bert-large (340M parameters). 
\section{Cases}
\label{sec:appendix4}
We illustrate cases of 8 different attribute combinations in Table \ref{tab:case}.

\clearpage
\onecolumn
\noindent \textcolor{dred}{WARNING: Next may contain contents that are offensive in nature.}
\begin{small}
\begin{longtable}{l|l|l}
          \hline 
          
          \hline
            \textbf{Attributes} & \textbf{Models} & \textbf{Generation Results}\\
            \hline
            \hline
            \multirow{18}{*}{\tabincell{l}{\textcolor{sred}{Neg.}\\ \textcolor{sblue}{World}\\ NonTox.}} 
            & \textbf{PPLM} & \tabincell{l}{\underline{Furthermore,} the government will be allowed to carry out airstrikes from inside \textcolor{sblue}{Afghanistan}\\ even after the end of \textcolor{sblue}{Nato}'s withdrawal, according to senior defence officials. \textcolor{sblue}{The UK} is also\\ planning to use air strikes \textcolor{sblue}{in Afghanistan and Iraq} from the skies over its bases.}\\
            \cline{2-3}
            & \textbf{GeDi} & \tabincell{l}{\underline{Furthermore,} the \textcolor{sblue}{Iraqi authorities} plan to seize petrol from \textcolor{sblue}{Shia} vehicles and trucks as well.\\ \textcolor{sblue}{Iraqi Foreign Minister Shaida al-Abadi} said \textcolor{sblue}{Iraq} was ready for such actions if \textcolor{sblue}{US President}\\ \textcolor{sblue}{Donald Trump} issued a military order ordering it.}\\
            \cline{2-3}
            & \tabincell{l}{\textbf{MU}\\\textbf{COCO}} & \tabincell{l}{\underline{Furthermore,} the new law will make it sale for any individual or entity to regulated keep give\\ away \textcolor{sblue}{Mumbai} firearm to anyone a a background check, including those who are prohibited\\ from doing so under \textcolor{sblue}{federal law}.}\\
            \cline{2-3}
            & \tabincell{l}{\textbf{Mix\&}\\\textbf{Match}} & \tabincell{l}{\underline{Furthermore,} the \textcolor{sblue}{uk} will be \textcolor{sred}{reluctant} to carry out airstrikes anywhere in \textcolor{sblue}{europe} until near\\ the end of \textcolor{sblue}{nato}'s mandate, according to some \textcolor{sblue}{nato} officials. the \textcolor{sblue}{uk} is likewise \textcolor{sred}{reluctant} to\\ drop warheads against \textcolor{sblue}{iran and iraq} from the air over \textcolor{sblue}{british territory} anywhere;}\\
            \cline{2-3}
            & \tabincell{l}{\textbf{Prefix}\\\textit{concate}} & \tabincell{l}{\underline{Furthermore,} the first and the first of his world. The world.S. The \textcolor{sblue}{U.S.} The U. The world's\\ country and a new-year.}\\
            \cline{2-3}
            & \tabincell{l}{\textbf{Prefix}\\\;\;\textit{semi}} & \tabincell{l}{\underline{Furthermore,} a new survey conducted by a new survey of the \textcolor{sblue}{Middle Eastern population} in the\\ country was revealed to be a very close match for the official record of the \textcolor{sblue}{National Socialist}\\ \textcolor{sblue}{Party (NTP)} in the country.}\\
            \cline{2-3}
            & \textbf{Ours} & \tabincell{l}{\underline{Furthermore,} the movie's main focus is \textcolor{sred}{getting rid of} \textcolor{sblue}{Robert Kennedy}. \textcolor{sred}{This movie has no plot,}\\ \textcolor{sred}{no action and no even remotely decent characterizations.} It's simply a \textcolor{sred}{glorified} version of \textcolor{sblue}{what}\\ \textcolor{sblue}{happened to George Bush in 2004}.}\\
            \hline
            \hline
            \multirow{18}{*}{\tabincell{l}{\textcolor{sred}{Neg.}\\\textcolor{sblue}{Sports}\\ NonTox.}} 
            & \textbf{PPLM} & \tabincell{l}{\underline{This essay} discusses the role of private security forces in Libya. The military's role in this \textcolor{sred}{crisis}\\ can be divided into two phases: 1) The first phase involved the transfer and transfer of the\\ control of the situation to a military body.}\\
            \cline{2-3}
            & \textbf{GeDi} & \tabincell{l}{\underline{This essay} discusses last season who was \textcolor{sred}{demoted away} from the \textcolor{sblue}{league} and how his \textcolor{sred}{decline}\\ in \textcolor{sblue}{playing time} impacted the \textcolor{sblue}{team} as a whole. With detailed observations, analysis, stories\\ provided by some of these \textcolor{sblue}{players including Orlando City fullback Ben Sweat} and Toronto.}\\
            \cline{2-3}
            & \tabincell{l}{\textbf{MU}\\\textbf{COCO}} & \tabincell{l}{\underline{This essay} discusses howwcsstore can Consent a more humane society and how Mold\\ willroximately the way webp topics our own Intake and our relationship with them.\\  enoughWhat is a body)? awa A body is the \sout{transsexual porn} Franch structureglers glucobos}\\
            \cline{2-3}
            & \tabincell{l}{\textbf{Mix\&}\\\textbf{Match}} & \tabincell{l}{\underline{This essay} from an official, who was investigating two suspected drug dealers, "\textcolor{sred}{failed} to find\\ any probable cause." he stated that "\textcolor{sred}{confusion reigned}" as the two men "\textcolor{sred}{struggled} for some\\ time" while evans "continued throwing punches."}\\
            \cline{2-3}
            & \tabincell{l}{\textbf{Prefix}\\\textit{concate}} & \tabincell{l}{\underline{This essay} discusses the fact of the original Germanic tradition of a man's attempt to make a\\ name on the \textcolor{sblue}{English football team} and the fact of the \textcolor{sblue}{English football league}.}\\
            \cline{2-3}
            &  \tabincell{l}{\textbf{Prefix}\\\;\;\textit{semi}} & \tabincell{l}{\underline{This essay} discusses the fact that the \textcolor{sblue}{NHL is not a national sport}. It also provides a new\\ perspective to the fact that the \textcolor{sblue}{NHL is not a major league}.}\\
            \cline{2-3}
            & \textbf{Ours}& \tabincell{l}{\underline{This essay} discusses how the \textcolor{sblue}{Miami Heat} \textcolor{sred}{lost} to the \textcolor{sblue}{Atlanta Hawks} in a \textcolor{sblue}{seven-minute overtime}\\last night, and \textcolor{sred}{how they should never again be able to make it} with their \textcolor{sblue}{team mates}.}\\
            
            \hline
            \hline
            \multirow{18}{*}{\tabincell{l}{\textcolor{sred}{Neg.}\\ \textcolor{sblue}{Business}\\ NonTox.}}
            & \textbf{PPLM} & \tabincell{l}{\underline{Foundational to this is} the need for a \textcolor{sblue}{national banking system} for the purpose of \textcolor{sblue}{financing}\\ \textcolor{sblue}{the banking system}. \textcolor{sblue}{The Federal Reserve} has already taken over this task by creating and\\ controlling the \textcolor{sblue}{money supply} in the form of the \textcolor{sblue}{Federal Reserve bank}, which is now owned\\ and operated by the Federal.}\\
            \cline{2-3}
            & \textbf{GeDi} & \tabincell{l}{\underline{Foundational to this is} the New York Int'l Fedal and \textcolor{sblue}{Foreign Market Team}. This practice\\ includes facilitating contacts between two levels of \textcolor{sblue}{financial institutions} as necessary for a\\ successful settlement of an equity \textcolor{sblue}{investment transaction}.}\\
            \cline{2-3}
            & \tabincell{l}{\textbf{MU}\\\textbf{COCO}} & \tabincell{l}{\underline{Foundational to this is} a RegulatorySPONSORED of community and \textcolor{sblue}{debtor}. not have a shared\\ history of \textcolor{sred}{Recession} \textcolor{sblue}{bourgeois} The struggles ofBuffpeople,SPONSORED the struggles of all\\ of our individuals bunk are interrelated,\_- we are all part ze the same struggle\_.}\\
            \cline{2-3}
            & \tabincell{l}{\textbf{Mix\&}\\\textbf{Match}} & \tabincell{l}{\underline{Foundational to this is} the woodward \textcolor{sblue}{foundation} is our belief that low - \textcolor{sblue}{income} housing has a\\ \textcolor{sred}{devastating} effect on mental health. our research indicates that \textcolor{sred}{depressive} symptoms are much\\ greater among the \textcolor{sblue}{economically} weaker. we believe that a depressed patient's life expectancy\\ will \textcolor{sred}{decline dramatically} as a result.}\\
            \cline{2-3}
            & \tabincell{l}{\textbf{Prefix}\\\textit{concate}} & \tabincell{l}{\underline{Foundational to this is} the \textcolor{blue}{company}.\textbackslash n\textbackslash n}\\
            \cline{2-3}
            & \tabincell{l}{\textbf{Prefix}\\\;\;\textit{semi}} & \tabincell{l}{\underline{Foundational to this is} the \textcolor{sblue}{company}'s ability to offer \textcolor{sblue}{a full range of services to customers}. The\\ \textcolor{sblue}{company} has been able to offer a good \textcolor{sblue}{product and service} for years.}\\
            \cline{2-3}
            & \textbf{Ours} & \tabincell{l}{\underline{Foundational to this is} \textcolor{sblue}{a \$1 million \textcolor{sred}{cut} from the company's annual budget}, which means it could\\ be \textcolor{sred}{difficult} to \textcolor{sblue}{cash out of its debt}.}\\
            
            \hline
            \hline
            \multirow{4}{*}{\tabincell{l}{\textcolor{sred}{Neg.}\\ \textcolor{sblue}{Sci/Tech}\\ NonTox.}} 
            & \textbf{PPLM} & \tabincell{l}{\underline{The last time} I wrote a \textcolor{sblue}{blog post} about the future of \textcolor{sblue}{Linux} and how to prepare for it I wrote: I\\ am very pleased to announce that the \textcolor{sblue}{Linux kernel} has been updated to 1.8.2! This is good\\ news for the \textcolor{sblue}{Linux}.}\\
            \cline{2-3}
            & \textbf{GeDi} & \tabincell{l}{\underline{The last time}  we examined this topic we estimated that 5\% of the age-appropriate weight was\\ unnecessary due to \textcolor{sblue}{red} fatherful \textcolor{sred}{neglect}; only 14\% of adopted \textcolor{sblue}{healthy}, fit barrels were\\ considered but maintained. Our \textcolor{sblue}{data} predict that at least 44lb.}\\
            \cline{2-3}
            & \tabincell{l}{\textbf{MU}\\\textbf{COCO}} & \tabincell{l}{\underline{The last time} I checked \textcolor{sblue}{VMware} would's \textcolor{sred}{not} like the world's largest and fastest ship was a \\single- \textcolor{sblue}{Anonlled organism} that could fly, and the last time I checked, it's not like the world's\\ largest and fastest ship the a single-}\\
            \cline{2-3}
            \multirow{5}{*}{\tabincell{l}{\textcolor{sred}{Neg.}\\ \textcolor{sblue}{Sci/Tech}\\ NonTox.}} 
            & \tabincell{l}{\textbf{Mix\&}\\\textbf{Match}} & \tabincell{l}{\underline{The last time} the lake was drained, however, to make way for a golf course at a nearby site,\\ and there were reports on a nearby island that a large-scale \textcolor{sblue}{groundwater cleaning program}\\ was beginning. the lake was drained, however, to make way elsewhere;}\\
            \cline{2-3}
            & \tabincell{l}{\textbf{Prefix}\\\textit{concate}} & \tabincell{l}{\underline{The last time} you are a fan, you have to be aware of the \textcolor{sred}{bad news} about the \textcolor{sblue}{Internet}. The\\ problem is that this isn't even remotely acceptable to the author.}\\
            \cline{2-3}
            & \tabincell{l}{\textbf{Prefix}\\\;\;\textit{semi}} & \tabincell{l}{\underline{The last time} I see this \textcolor{sblue}{film} I have not seen it in my life. I have watched it in a very limited\\ number of days and I have been \textcolor{sred}{very disappointed}. The acting is \textcolor{sred}{terrible} and the acting is \textcolor{sred}{bad}.}\\
            \cline{2-3}
            & \textbf{Ours} & \tabincell{l}{\underline{The last time} I saw this \textcolor{sblue}{film} was in the theater. It was \textcolor{sred}{terribly disappointing}. There is no plot or\\suspense whatsoever, nor any action whatsoever. The only thing that can be attributed to this\\ movie is \textcolor{sred}{the lack of} \textcolor{sblue}{a sound system}.}\\
            
            \hline
            \hline
            \multirow{18}{*}{\tabincell{l}{\textcolor{sred}{Pos.}\\ \textcolor{sblue}{World}\\ NonTox.}} 
            & \textbf{PPLM} & \tabincell{l}{\underline{The connection} between obesity and autism has been identified for the first time using a unique\\ antibody screening test, according to researchers with the University of Texas MD Anderson\\ Cancer Center. They found that a protein called CD34 has a \textcolor{sred}{powerful impact} on autism.}\\
            \cline{2-3}
            & \textbf{GeDi} & \tabincell{l}{\underline{The connection} between \textcolor{sblue}{Greece and Russia} reached new heights through cooperation on a\\ number of initiatives States Parties undertook joint action to crack down on corruption abroad.\\ For instance, the \textcolor{sblue}{Russian Federation} launched an all-cash inquiry aimed at identifying persons.}\\
            \cline{2-3}
            & \tabincell{l}{\textbf{MU}\\\textbf{COCO}} & \tabincell{l}{\underline{The connection} between staking two is not a loneliness of mere coincidence. The connection\\ snowball dividing matter of history, and history BW a history arresting its own, of which hero\\ are all the victims analyse nogyI don't believe in coincidence Alger said the}\\
            \cline{2-3}
            & \tabincell{l}{\textbf{Mix\&}\\\textbf{Match}} & \tabincell{l}{\underline{The connection} is an illustration of the moon, from the book 'the lord of light and darkness',\\ by william shakespeare ( photo courtesy of william shakespeare ). an illustration of the sun,\\ from the book 'the lord of light and darkness', by william shakespeare (photo courtesy)}\\
            \cline{2-3}
            & \tabincell{l}{\textbf{Prefix}\\\textit{concate}} & \tabincell{l}{\underline{The connection} of the \textcolor{sblue}{United States}'s the world of the world's first-year of a new-run of the\\ world in the world.S.}\\
            \cline{2-3}
            & \tabincell{l}{\textbf{Prefix}\\\;\;\textit{semi}} & \tabincell{l}{\underline{The connection} between the world of the \textcolor{sblue}{American National Rifle Association} and the \textcolor{sblue}{United}\\ \textcolor{sblue}{States} is \textcolor{sred}{a fascinating, fascinating, and hilarious tale}. It has been an \textcolor{sred}{honor} to see the film on the\\ National Library shelves, and I am \textcolor{sred}{proud} to see the film.}\\
            \cline{2-3}
            & \textbf{Ours} & \tabincell{l}{\underline{The connection} between John Lennon and the \textcolor{sblue}{United States} is \textcolor{sred}{as strong as ever}. The Los\\ Angeles Times reports that Lee Sternberg's performance of his song "Lenny Luerer" \textcolor{sred}{won a}\\ \textcolor{sred}{round of applause} in \textcolor{sblue}{the U.S. Senate}.}\\

            \hline
            \hline
            \multirow{18}{*}{\tabincell{l}{ \textcolor{sred}{Pos.}\\ \textcolor{sblue}{Sports}\\ NonTox.}} 
            & \textbf{PPLM} & \tabincell{l}{\underline{More importantly,} the first day of the \textcolor{sblue}{2017 NFL Draft is always} \textcolor{sred}{exciting} to watch with fans,\\ because the \textcolor{sblue}{league} is going to get a lot of \textcolor{sred}{great talent} on defense in the coming weeks. The\\ \textcolor{sred}{biggest} draft prospect to come out this year, Alabama DT Vic Beasley}\\
            \cline{2-3}
            & \textbf{GeDi} & \tabincell{l}{\underline{More importantly,} I \textcolor{sred}{appreciated his honesty} along the way. \textcolor{sblue}{Orlando Pace} is usually a shadow\\ of his former self, but he understood the importance of all that went into this \textcolor{sred}{win and smiled}\\ \textcolor{sred}{again}.}\\
            \cline{2-3}
            & \tabincell{l}{\textbf{MU}\\\textbf{COCO}} & \tabincell{l}{\underline{More importantly,} he was able to defenders it work. it two men Fans in likeness fans on a\\ \textcolor{sblue}{Rugby coach}. He had justovich from medical trip that Europe and was looking partners a\\ place to eat. loved had never been in a bar}\\
            \cline{2-3}
            & \tabincell{l}{\textbf{Mix\&}\\\textbf{Match}} & \tabincell{l}{\underline{More importantly,} the sixth game of the \textcolor{sblue}{2018 stanley cup finals} presents a new challenge\\ and an \textcolor{sred}{exciting new opportunity}. the philadelphia flyers and pittsburgh penguins are joining\\ forces for a \textcolor{sblue}{six-game road trip} that begins in the nation's capital each weekend.}\\
            \cline{2-3}
            & \tabincell{l}{\textbf{Prefix}\\\textit{concate}} & \tabincell{l}{\underline{More importantly,} I have to remind everyone that this is a real story, so the fact that the two men\\ were not a couple of people who have to be treated as one of those who would be involved with\\ the \textcolor{sblue}{team}.}\\
            \cline{2-3}
            & \tabincell{l}{\textbf{Prefix}\\\;\;\textit{semi}} & \tabincell{l}{\underline{More importantly,} the \textcolor{sblue}{Boston Red Sox} have lost \textcolor{sblue}{the league title}, and the \textcolor{sblue}{players} themselves are\\ not yet qualified to be the best \textcolor{sblue}{player in the league}. The fact that they are not even qualified to\\ \textcolor{sblue}{play a match} of the best.}\\
            \cline{2-3}
            & \textbf{Ours} & \tabincell{l}{\underline{More importantly,} the \textcolor{sblue}{Houston Astros} \textcolor{sred}{won a great opportunity} to make a comeback with a\\ \textcolor{sred}{victory} over the Detroit Tigers in the \textcolor{sblue}{National League West}. The \textcolor{sblue}{team} has an \textcolor{sred}{outstanding}\\ \textcolor{sblue}{offensive line} and is \textcolor{sred}{tied for fifth in scoring among the nation}.}\\
            
            \hline
            \hline
            \multirow{12}{*}{\tabincell{l}{\textcolor{sred}{Pos.}\\ \textcolor{sblue}{Business}\\ NonTox.}} 
            & \textbf{PPLM} & \tabincell{l}{\underline{In brief,} the \textcolor{sblue}{federal tax law} allows \textcolor{sblue}{employers} to deduct up to 20\% of \textcolor{sblue}{compensation expenses}\\ from workers' \textcolor{sblue}{paychecks}. This deduction is a big deal because many \textcolor{sblue}{employees} have to pay\\ high \textcolor{sblue}{deductibles} for medical care.}\\
            \cline{2-3}
            & \textbf{GeDi} & \tabincell{l}{\underline{In brief,} Heiltsuk said that she holds central, shared concerns regarding how First Nations youth\\ can navigate \textcolor{sblue}{financial} injustices faced by society and why net aboriginal \textcolor{sblue}{debt} was surpassed in\\ 2015. All eight First Nations elected delegates at Monday's meeting}\\
            \cline{2-3}
            & \tabincell{l}{\textbf{MU}\\\textbf{COCO}} & \tabincell{l}{\underline{In brief,} Bach "sus anthologyions pione excel the outstanding Russian Returns in the hacking\\ of \textcolor{sblue}{capitalists economics} Committee letters were not whirlwind. But that's not what the White\\ Airways said in statement Alibaba late Tuesday afternoonisSpecial \textcolor{sblue}{Orderable}}\\
            \cline{2-3}
            & \tabincell{l}{\textbf{Mix\&}\\\textbf{Match}} & \tabincell{l}{\underline{In brief,} the u. s. department of agriculture ( usda ) produced a comprehensive list of how many\\ \textcolor{sblue}{jobs} were created in 2016, it identified 3. 1 million \textcolor{sblue}{jobs in the agriculture sector}, a \textcolor{sred}{dramatic}\\ \textcolor{sred}{uplift} from 2015's 2 million.}\\
            \cline{2-3}
            & \tabincell{l}{\textbf{Prefix}\\\textit{concate}} & \tabincell{l}{\underline{In brief,} the new of its \textcolor{sblue}{company}.\textbackslash n\textbackslash n}\\

            \cline{2-3}
            \multirow{2}{*}{\tabincell{l}{\textcolor{sred}{Pos.}\\ \textcolor{sblue}{Business}\\ NonTox.}} 
            & \tabincell{l}{\textbf{Prefix}\\\;\;\textit{semi}} & \tabincell{l}{\underline{In brief,} it is the \textcolor{sred}{best} movie I have ever seen, and \textcolor{sred}{I love it}. The movie is a \textcolor{sred}{perfect blend of}\\ \textcolor{sred}{comedy and comedy}. It is not a classic movie, but it is not a great movie.}\\
            \cline{2-3}
            & \textbf{Ours} & \tabincell{l}{\underline{In brief,} the \textcolor{sblue}{economy} is surged in July, \textcolor{sred}{boosted} by \textcolor{sred}{strong} \textcolor{sblue}{sales of oil and other products}, as well\\ as \textcolor{sred}{strong growth} in \textcolor{sblue}{U.S. manufacturing}.}\\
            
            \hline
            \hline
            \multirow{18}{*}{\tabincell{l}{\textcolor{sred}{Pos.}\\ \textcolor{sblue}{Sci/Tech}\\ NonTox.}} 
            & \textbf{PPLM} & \tabincell{l}{\underline{The country}'s first \textcolor{sblue}{solar power system}, built by a group of students at Harvard University, is\\ now operating. The project is aimed at encouraging \textcolor{sblue}{solar energy} development by encouraging\\ collaboration among universities, community groups and individuals.}\\
            \cline{2-3}
            & \textbf{GeDi} & \tabincell{l}{\underline{The country} illustrated \textcolor{sred}{beautifully} reflects the complexity of lives and customs.}\\
            \cline{2-3}
            & \tabincell{l}{\textbf{MU}\\\textbf{COCO}} & \tabincell{l}{\underline{The country}'s top diplomat, Blockchain Lavrov \textcolor{sblue}{IBM} said the UydiaS. was "very much looking\\ into" the matter. \textcolor{sred}{pleasantly} \textcolor{sblue}{engineers} Rapp a Bridges supplier of\textcolor{sblue}{hacker} vegan Iran, has been\\ trying to improve ties with \textcolor{sblue}{blockchain}, a \textcolor{sred}{close ally} and}\\
            \cline{2-3}
            & \tabincell{l}{\textbf{Mix\&}\\\textbf{Match}} & \tabincell{l}{\underline{The country} focuses on the role the united states has played in discovering new \textcolor{sblue}{technologies}\\ for the \textcolor{sblue}{advancement of science}, according to two u. s. officials briefed on - site. both officials,\\ newly appointed to handle national security matters \textcolor{sred}{welcomed} the sensitive nature of the\\ investigation.}\\
            \cline{2-3}
            & \tabincell{l}{\textbf{Prefix}\\\textit{concate}} & \tabincell{l}{\underline{The country}'s top \textcolor{sblue}{TV} channel is now a very \textcolor{sred}{popular} \textcolor{sblue}{TV} show. The only thing is the name.\\ I'm sure there are many people who would be willing to take it seriously, but I'll be \sout{damned} to\\ find out if they have a lot}\\
            \cline{2-3}
            & \tabincell{l}{\textbf{Prefix}\\\;\;\textit{semi}} & \tabincell{l}{\underline{The country}'s most famous \textcolor{sblue}{TV} series is the \textcolor{sred}{best and most powerful} show ever made. The story\\ is \textcolor{sred}{great}, the action is \textcolor{sred}{good}, the plot is \textcolor{sred}{great}, and the story is \textcolor{sred}{very good}. The cast is \textcolor{sred}{great}.}\\
            \cline{2-3}
            & \textbf{Ours} & \tabincell{l}{\underline{The country}'s \textcolor{sred}{biggest} \textcolor{sblue}{television network} has announced that it will offer a new version of the\\ \textcolor{sblue}{movie} which is based upon the \textcolor{sred}{popular} "Star Trek" series. It's \textcolor{sred}{truly amazing} to see how many\\ people are involved in making this movie so far.}\\

          \hline

          \hline
\caption{Generated Cases. \textcolor{sred}{Red} highlights the sentiment-related content. \textcolor{sblue}{Blue} highlights the topic-related content. \underline{Underlined} are the input prompts. \sout{Strikethrough} indicates toxic content.}
  \label{tab:case}
\end{longtable}
\end{small}
\twocolumn

\onecolumn
\section{Detailed Results}
\label{sec:appendix5}
\begin{small}
\begin{longtable}{l|cc|cccc|c}
          \hline 
          
          \hline
          \multirow{2}{*}{\textbf{Methods}} & \multicolumn{2}{c|}{\textbf{Sentiment} (\%)} &\multicolumn{4}{c|}{\textbf{Topic} (\%)} & \multirow{2}{*}{\textbf{Detox.} (\%)}\\
           & \textbf{Neg.}& \textbf{Pos.}& \textbf{World}&\textbf{Sports}& \textbf{Business}&\textbf{Sci./Tech.}&\\
          \hline
          \hline
          \multicolumn{6}{l}{\quad \textit{Weighted Decoding Based Methods}}\\
          \hline
          \textbf{PPLM} \textit{single-aspect} & 97.2& 62.7 & 74.9  & 46.5 & 62.4 & 98.6 & 93.2 \\
          \hline
          \multirow{8}{*}{\textbf{PPLM}}
          & 92.2 & - & 75.4 & - & - & - & 82.0\\
          & 84.4 & - & - & 41.8 & - & - & 76.0\\
          & 87.5 & - & - & - & 61.5 & - & 82.9\\
          & 85.3 & - & - & - & - & 95.0 & 76.2\\
          & - & 35.4 & 59.1 & - & - & - & 90.4\\
          & - & 39.5 & - & 34.1 & - & - & 89.5\\
          & - & 40.9 & - & - & 48.3 & - & 91.2\\
          & - & 52.7 & - & - & - & 93.1 & 91.3\\
          \cline{2-8}
          \qquad \textit{average} & 87.4 & 42.1 & 67.3 & 38.0 & 54.9 & 94.1 & 84.9\\
          \hline
          \hline
          \textbf{GeDi} \textit{single-aspect} & 93.9& 70.7& 73.4  & 85.7 & 75.7 & 98.0 & 94.9 \\
          \hline
          \multirow{8}{*}{\textbf{GeDi}}
          & 94.7 & - & 80.0 & - & - & - & 90.6\\
          & 84.2 & - & - & 74.8 & - & - & 93.9\\
          & 94.9 & - & - & - & 75.7 & - & 96.6\\
          & 90.6 & - & - & - & - & 80.1 & 92.8\\
          & - & 53.7 & 61.4 & - & - & - & 94.4\\
          & - & 60.5 & - & 74.3 & - & - & 95.2\\
          & - & 57.6 & - & - & 54.3 & - & 95.7\\
          & - & 72.3 & - & - & - & 90.2 & 94.2\\
          \cline{2-8}
          \qquad \textit{average} & 91.1 & 61.0 & 70.7 & 74.6 & 65.0 & 85.2 & 94.2\\
          \hline
          \hline
          \multicolumn{6}{l}{\quad \textit{Multi-Objective Optimization Based Methods}}\\
          \hline
          \multirow{8}{*}{\textbf{MUCOCO}}
          & 97.9 & - & 54.5 & - & - & - & 85.7\\
          & 94.6 & - & - & 55.8 & - & - & 85.7\\
          & 96.8 & - & - & - & 65.6 & - & 87.3\\
          & 95.5 & - & - & - & - & 96.1 & 86.9\\
          & - & 30.4 & 48.0 & - & - & - & 91.0\\
          & - & 26.3 & - & 59.8 & - & - & 92.6\\
          & - & 34.6 & - & - & 62.1 & - & 93.8\\
          & - & 43.9 & - & - & - & 95.1 & 93.1\\
          \cline{2-8}
          \qquad \textit{average} & 96.2 & 33.8 & 51.3 & 57.8 & 63.9 & 95.6 & 89.5\\
          \hline
          \hline
          \textbf{Mix\&Match} \textit{single-aspect} & 99.2 & 63.3 & 79.5  & 57.4 & 69.6 & 99.3 & 96.9 \\
          \hline
          \multirow{8}{*}{\textbf{Mix\&Match}}
          & 96.1 & - & \textbf{80.6} & - & - & - & 93.1\\
          & 97.7 & - & - & 48.2 & - & - & 93.0\\
          & 98.2 & - & - & - & 66.6 & - & 97.0\\
          & 96.8 & - & - & - & - & 99.6 & 96.1\\
          & - & 53.0 & 67.3 & - & - & - & 95.5\\
          & - & 45.0 & - & 44.0 & - & - & 96.7\\
          & - & 41.5 & - & - & 55.8 & - & 97.7\\
          & - & 59.7 & - & - & - & 97.3 & 97.5\\
          \cline{2-8}
          \qquad \textit{average} & \textbf{97.2} & 49.8 & \textbf{74.0} & 46.1 & 61.2 & 98.5 & 95.8 \\
          \hline
          \hline
          \multicolumn{6}{l}{\quad \textit{Prefix-Tuning Based Methods}}\\
          \hline
          \textbf{Prefix} \textit{single-aspect} & 88.4 & 90.6 & 74.5 & 85.3 & 93.5 & 93.6 & 93.8\\
          \hline
          \hline
          \multirow{8}{*}{\tabincell{l}{\textbf{Contrastive Prefix}\\ \quad concatenation}} 
          & 32.4 & - & 50.3 & - & - & - & 90.9\\
          & 88.1 & - & - & 73.8 & - & - & 89.1\\
          & 51.6 & - & - & - & 70.0 & - & 94.1\\
          & 94.3 & - & - & - & - & 94.1 & 88.3\\
          & - & 77.6 & 46.8 & - & - & - & 92.2\\
          & - & 70.2 & - & 78.5 & - & - & 95.9\\
          & - & 51.9 & - & - & 73.1 & - & 94.7\\
          & - & 72.0 & - & - & - & 88.1 & 95.6\\
          \cline{2-8}
          \qquad \textit{average} & 66.6 & 67.9 & 48.5 & 76.2 & 71.6 & 91.1 & 92.6\\
          \hline
          \multirow{8}{*}{\tabincell{l}{\textbf{Contrastive Prefix}\\ \quad semi-supervised}}
          & 65.5 & - & 80.6 & - & - & - & 91.8\\
          & 67.2 & - & - & 90.3 & - & - & 92.5\\
          & 56.0 & - & - & - & 79.2 & - & 92.2\\
          & 90.0 & - & - & - & - & 93.3 & 84.8\\
          & - & 93.5 & 64.8 & - & - & - & 95.1\\
          & - & 41.8 & - & 78.5 & - & - & 94.8\\
          & - & 87.4 & - & - & 41.7 & - & 95.2\\
          & - & 93.6 & - & - & - & 86.7 & 95.3\\
          \cline{2-8}
          \qquad \textit{average} & 69.7 & 79.1 & 72.7 & \textbf{84.4} & 60.5 & 90.0 & 92.7\\
          
          \hline
          \hline

          \multirow{8}{*}{\textbf{Ours}}
          & 69.7 & - & 71.7 & - & - & - & 84.1\\
          & 78.6 & - & - & 80.0 & - & - & 80.2\\
          & \textbf{99.9} & - & - & - & \textbf{96.7} & - & 96.8\\
          & 92.8 & - & - & - & - & 98.0 & 81.7\\
          & - & 80.5 & 58.0 & - & - & - & 95.1\\
          & - & 84.7 & - & 86.6 & - & - & 94.5\\
          & - & 87.6 & - & - & 91.7 & - & \textbf{98.1}\\
          & - & \textbf{99.7} & - & - & - & 96.1 & 95.4\\
          \cline{2-8}
          \qquad \textit{average} & 85.3 & \textbf{88.1} & 64.9 & 83.3 & \textbf{94.2} & 96.8 & 90.7\\
          
          \hline
          
          \multirow{8}{*}{\tabincell{l}{\textbf{Ours}\\ \quad w/o Asp. Loss}}
          & 64.3 & - & 51.8 & - & - & - & 90.1\\
          & 71.5 & - & - & 71.0 & - & - & 93.4\\
          & 68.2 & - & - & - & 59.9 & - & 95.7\\
          & 62.4 & - & - & - & - & \textbf{99.8} & 96.0\\
          & - & 92.0 & 60.6 & - & - & - & 97.6\\
          & - & 59.4 & - & \textbf{93.8} & - & - & 94.3\\
          & - & 86.8 & - & - & 72.1 & - & 97.9\\
          & - & 68.3 & - & - & - & 98.4 & 97.2\\
          \cline{2-8}
          \qquad \textit{average} & 66.6 & 76.6 & 56.2 & 82.4 & 66.0 & \textbf{99.1} & 95.3\\
          
          \hline
          
          \multirow{8}{*}{\tabincell{l}{\textbf{Ours}\\ \quad w/o Att. Loss}}
          & 99.2 & - & 15.2 & - & - & - & 96.5\\
          & 99.8 & - & - & 36.5 & - & - & 96.3\\
          & 97.8 & - & - & - & 17.9 & - & 95.4\\
          & 84.9 & - & - & - & - & 97.7 & 95.6\\
          & - & \; 3.2 & 14.4 & - & - & - & 96.3\\
          & - & \; 0.1 & - & 40.4 & - & - & 96.0\\
          & - & \; 1.3 & - & - & 13.9 & - & 95.7\\
          & - & \; 6.5 & - & - & - & 97.7 & 95.8\\
          \cline{2-8}
          \qquad \textit{average} & 95.4 & \; 5.6 & 14.8 & 38.5 & 15.9 & 97.7 & \textbf{96.0}\\

          \hline

          \hline

\caption{Detailed Combination Results on Multi-Aspect Control. }
  \label{apptab:1}
\end{longtable}
\end{small}
\twocolumn

\end{document}